%% file: main.tex
\documentclass[sigconf]{acmart}
\usepackage{multirow}
\usepackage{graphicx}
\usepackage{subfigure}
\usepackage[normalem]{ulem}
\usepackage{url}
\usepackage{longtable}
\usepackage{enumerate}
\usepackage{booktabs}
\usepackage{multirow}
\usepackage{balance}
\usepackage{listings}
\usepackage{xspace}

\usepackage{amsmath}
\usepackage{supertabular}
\usepackage{tcolorbox}
\usepackage{colortbl}
\usepackage{color}
\usepackage{threeparttable}
\usepackage{ulem}
\usepackage{enumitem}
\usepackage{amssymb}
\usepackage{pifont}

\usepackage{float}
\usepackage[ruled,linesnumbered]{algorithm2e}
\usepackage{algorithm2e, setspace, algpseudocode}

\newcommand{\ie}{\textit{i.e.,}\xspace}

\newtheorem{definition}{Definition}

\definecolor{MyRed}{HTML}{a22f2c}
\definecolor{MyRed2}{HTML}{e22046}
\definecolor{MyYellow}{HTML}{9e7132}
\definecolor{MyGreen}{HTML}{387f61}
\definecolor{MyBlue}{HTML}{2278BF} 
\definecolor{fbApp}{HTML}{ffe4e3}
\definecolor{tabhighlight}{HTML}{e5e5e5}
\newcommand{\rowc}{\rowcolor{fbApp}}

\AtBeginDocument{%
  }

\copyrightyear{2025}
\acmYear{2025}
\setcopyright{rightsretained}

\copyrightyear{2025}
\acmYear{2025}
\setcopyright{cc}
\setcctype{by}
\acmConference[KDD '25]{Proceedings of the 31st ACM SIGKDD Conference on
Knowledge Discovery and Data Mining V.2}{August 3--7, 2025}{Toronto, ON,
Canada}
\acmBooktitle{Proceedings of the 31st ACM SIGKDD Conference on Knowledge
Discovery and Data Mining V.2 (KDD '25), August 3--7, 2025, Toronto, ON,
Canada}
\acmDOI{10.1145/3711896.3736860}
\acmISBN{979-8-4007-1454-2/2025/08}

\makeatother

\begin{document}

\title{BLAST: Balanced Sampling Time Series Corpus for Universal Forecasting Models}

\author{Zezhi Shao}
\authornote{Equal contribution.}
\affiliation{
\institution{Institute of Computing Technology, \\Chinese Academy of Sciences}
\institution{State Key Laboratory of AI Safety}
\institution{University of Chinese Academy of Sciences}
\country{}
}
\email{shaozezhi@ict.ac.cn}

\author{Yujie Li}
\authornotemark[1]
\affiliation{
\institution{Institute of Computing Technology, \\Chinese Academy of Sciences}
\institution{State Key Laboratory of AI Safety}
\institution{University of Chinese Academy of Sciences}
\country{}
}
\email{liyujie23s@ict.ac.cn}

\author{Fei Wang}
\authornote{Corresponding author.}
\affiliation{
\institution{Institute of Computing Technology, \\Chinese Academy of Sciences}
\institution{State Key Laboratory of AI Safety}
\institution{University of Chinese Academy of Sciences}
\country{}
}
\email{wangfei@ict.ac.cn}

\author{Chengqing Yu, Yisong Fu}
\affiliation{
\institution{Institute of Computing Technology, \\Chinese Academy of Sciences}
\institution{State Key Laboratory of AI Safety}
\institution{University of Chinese Academy of Sciences}
\country{}
}
\email{{yuchengqing22b, fuyisong24s}@ict.ac.cn}

\author{Tangwen Qian, Bin Xu, Boyu Diao}
\affiliation{
\institution{Institute of Computing Technology, \\Chinese Academy of Sciences}
\institution{State Key Laboratory of AI Safety}
\institution{University of Chinese Academy of Sciences}
\country{}
}
\email{{qiantangwen, xubin, diaoboyu2012}@ict.ac.cn}

\author{Yongjun Xu, Xueqi Cheng}
\affiliation{
\institution{Institute of Computing Technology, \\Chinese Academy of Sciences}
\institution{State Key Laboratory of AI Safety}
\institution{University of Chinese Academy of Sciences}
\country{}
}
\email{{xyj,cxq}@ict.ac.cn}

\renewcommand{\shortauthors}{Zezhi Shao et al.}

\begin{abstract}
The advent of universal time series forecasting models has revolutionized zero-shot forecasting across diverse domains, yet the critical role of data diversity in training these models remains underexplored. Existing large-scale time series datasets often suffer from inherent biases and imbalanced distributions, leading to suboptimal model performance and generalization. To address this gap, we introduce BLAST, a novel pre-training corpus designed to enhance data diversity through a balanced sampling strategy. First, BLAST incorporates 321 billion observations from publicly available datasets and employs a comprehensive suite of statistical metrics to characterize time series patterns. Then, to facilitate pattern-oriented sampling, the data is implicitly clustered using grid-based partitioning. Furthermore, by integrating grid sampling and grid mixup techniques, BLAST ensures a balanced and representative coverage of diverse patterns. Experimental results demonstrate that models pre-trained on BLAST achieve state-of-the-art performance with a fraction of the computational resources and training tokens required by existing methods. Our findings highlight the pivotal role of data diversity in improving both training efficiency and model performance for the universal forecasting task.

\end{abstract}



\vspace{-0.2cm}

\maketitle

\vspace{-0.2cm}

\newcommand\kddavailabilityurl{https://github.com/GestaltCogTeam/BasicTS}

\ifdefempty{\kddavailabilityurl}{}{
\begingroup\small\noindent\raggedright\textbf{KDD Availability Link:}\\
The code for training universal forecasting models with BLAST is available at {\url{https://github.com/GestaltCogTeam/BasicTS}}, and the BLAST generation code can be found at {\url{https://github.com/GestaltCogTeam/BLAST}}.
\endgroup
}

\vspace{-0.2cm}
\input{sections/introduction}

\input{sections/preliminaries}
\input{sections/related_work}

\input{sections/method}
\input{sections/experiments}

\input{sections/conclusion}

\bibliographystyle{ACM-Reference-Format}
\balance
\bibliography{references}

\input{sections/appendix}

\end{document}

%% file: sections/introduction.tex
\section{Introduction}
\label{sec_1}

\begin{figure}[t]
  \centering
  \setlength{\abovecaptionskip}{0.1cm}
  \setlength{\belowcaptionskip}{-0.2cm}
  \includegraphics[width=0.95\linewidth]{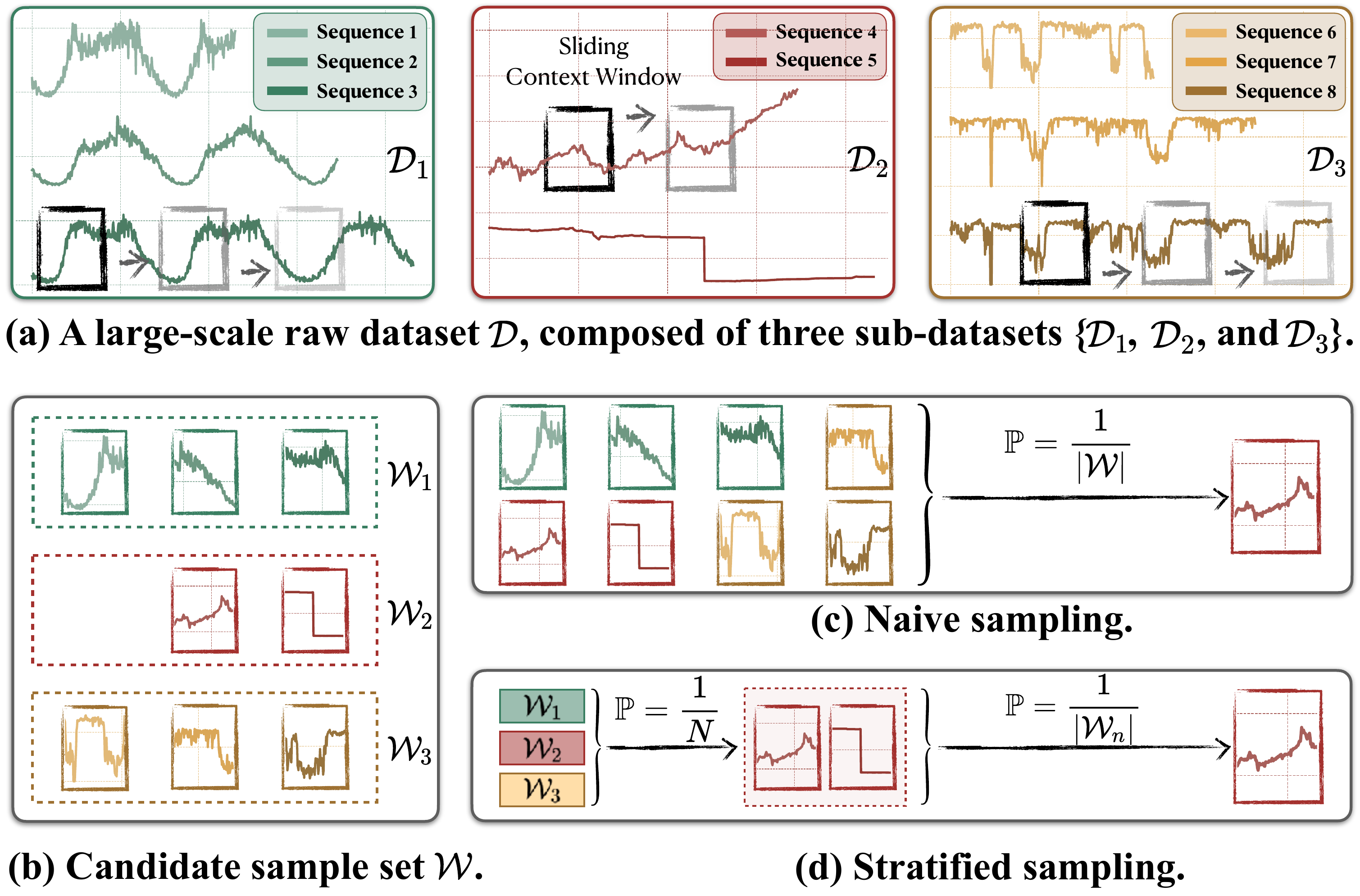}
  \caption{Illustration of the large-scale time series forecasting pre-training dataset and various sampling methods.}
  \label{Intro1}
\end{figure}

Universal time series forecasting models have introduced new possibilities for accurate zero-shot forecasting across various domains~\cite{LagLLaMA, TimeGPT1, Chronos, TimesFM, MOIRAI, TimeMoE, Timer}.
One of the most critical foundations for training these models lies in large-scale and diverse datasets.
Consequently, acquiring and organizing these training corpora has emerged as a crucial challenge.

A large-scale time series dataset is typically composed of multiple sub-datasets, where candidate samples are generated using a sliding window on each sequence and subsequently sampled to obtain data for model training.
An example of a large-scale dataset consisting of three sub-datasets is illustrated in Figure \ref{Intro1}(a).
It is worth noting that the sequence length and the number of sequences may vary significantly across sub-datasets.
Recent pioneering studies~\cite{Chronos,MOIRAI,Timer,TimeMoE,ForecastPFN} have leveraged multi-domain data to construct large-scale time series datasets.
For instance, the LOTSA~\cite{MOIRAI} dataset contains over 231 billion observations (considering all variates), while the Time-300B~\cite{TimeMoE} dataset is even larger, with 309 billion observations.
These studies primarily focus on the \textit{scale} of data, laying a solid foundation for training universal forecasting models.

\begin{figure}[t]
  \centering
  \setlength{\abovecaptionskip}{0.1cm}
  \setlength{\belowcaptionskip}{-0.2cm}
  \includegraphics[width=0.95\linewidth]{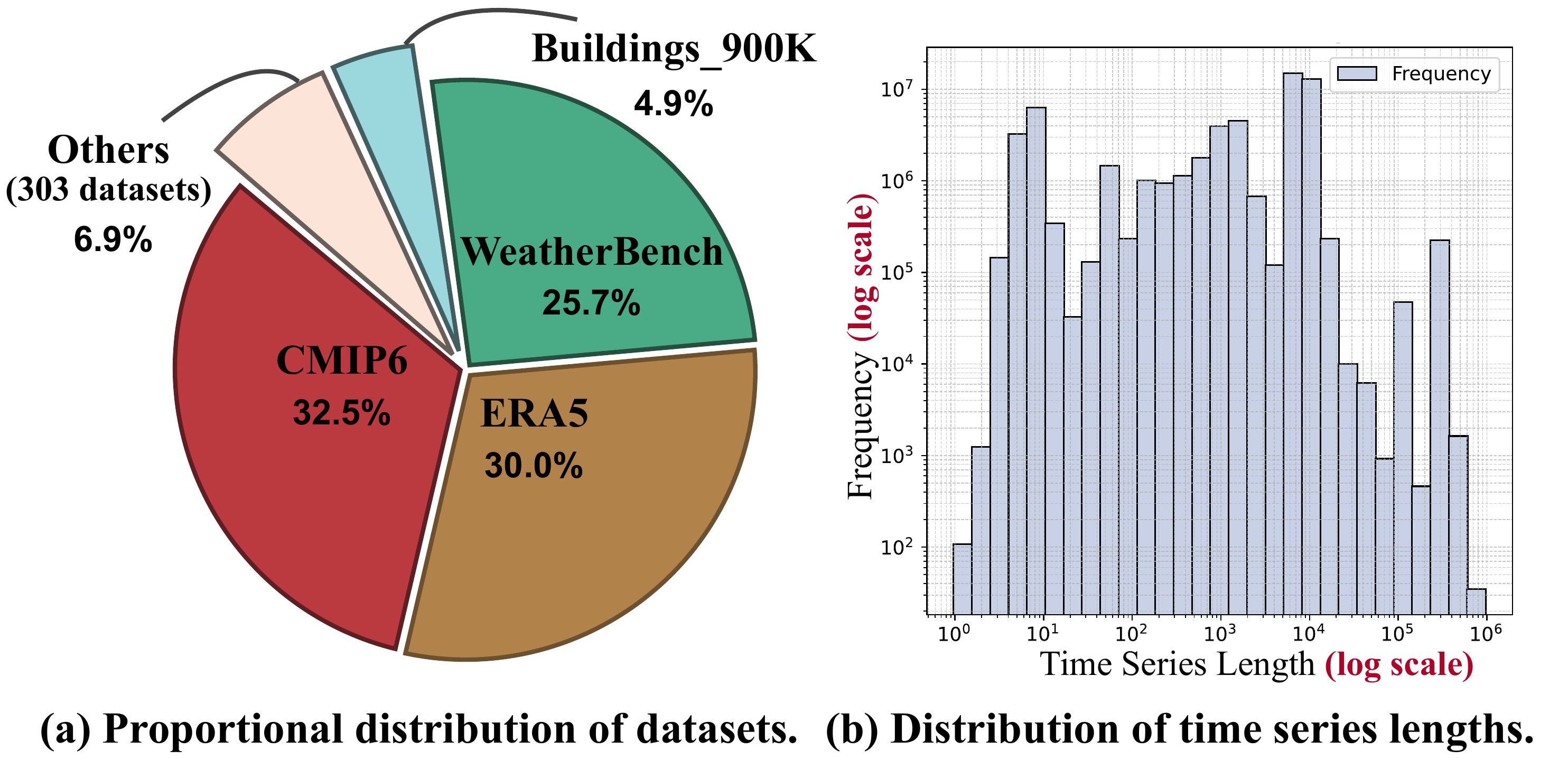}
  \caption{The uneven distribution of the raw large-scale time series dataset collected by BLAST.}
  \label{Intro2}
\end{figure}

Despite the growing scale, the \textit{diversity} of pre-training data has not yet been investigated.
High‐quality training data should capture a wide array of patterns while ensuring balanced sample sizes for each~\cite{SemDeDup,ClusterClip,miao2024less}.
However, the initial distribution of large-scale time series datasets is often highly imbalanced.
As illustrated in Figure \ref{Intro2}(a), only three datasets account for 88.2\% of the total data volume, and Figure~\ref{Intro2}(b) further highlights the imbalance in sequence lengths, where longer sequences tend to contribute disproportionately more samples.
These skewed distributions will result in numerous repetitive patterns~\cite{ChatTime, TimeMoE} in the raw data, compromising overall data diversity.
Thus, how to sample data with rich and balanced patterns becomes a crucial challenge.

However, existing studies generally overlook these imbalance issues, adopting simplistic sampling strategies such as naive sampling or stratified sampling. 
The former uniformly selects samples from all sub-datasets, as shown in Figure~\ref{Intro1}(c).
The latter usually involves two steps: first, uniformly or weightedly selecting a sub-dataset (or sub-domain), and then selecting a sample within that sub-dataset, as illustrated in Figure~\ref{Intro1}(d).
While these sampling strategies are intuitive and easy to implement, they fail to sufficiently correct for the inherent biases in large‐scale time series data.
Specifically, while naive sampling entirely overlooks these biases, stratified sampling attempts to mitigate them, but often assumes that data within the same dataset or domain share similar patterns, which is reasonable but does not always hold.
For instance, as shown in Figure \ref{Intro1}(a), both $\mathcal{D}_1$ and $\mathcal{D}_3$ originate from the traffic domain but exhibit distinct patterns. Similarly, two time series within $\mathcal{D}_2$ display divergent patterns. 
In summary, the inability to ensure diversity in the training data can have significant negative consequences.
For example, the model may overfit to frequent patterns while underfitting less common ones, impairing its generalization capability.

To address the aforementioned issues, we propose a novel pre-training corpus named BLAST~(\underline{B}a\underline{LA}nced \underline{S}ampling \underline{T}ime series corpus).
First, we integrate a wide range of publicly available datasets, creating a large-scale dataset with a total of 321 billion observations.
Unlike prior approaches that depend on dataset or domain labels to differentiate time series patterns, BLAST incorporates a diverse array of statistical attributes to comprehensively characterize each time series' patterns, such as stationarity, seasonality, volatility, \textit{etc}.
Subsequently, BLAST amalgamates these heterogeneous features into unified feature vectors through a discretization process and projects them into a low-dimensional space, thereby intuitively revealing the uneven distribution of the data.
Then, BLAST employs grid sampling and grid mixup within the low-dimensional space to ensure a balanced and representative coverage of diverse patterns.

To validate the effectiveness of BLAST, we trained state-of-the-art universal forecasting models using the proposed corpus \textit{from scratch}.
Table~\ref{tab:Intro1} presents the results from the TimeMoE~\cite{TimeMoE} model.\footnote{The choice of TimeMoE as the baseline for presenting the results is motivated by two main reasons: (i) TimeMoE is the only model pre-trained on a comparably large-scale raw dataset (309 billion observations); and (ii) TimeMoE is recognized as one of the state-of-the-art universal forecasting models.}
The original TimeMoE was trained on \textit{419 billion tokens} using \textit{128 A100 GPUs}.
In contrast, the BLAST-based TimeMoE achieves state-of-the-art performance with only \textit{78 billion tokens} and \textit{8 A100 GPUs}.
These results demonstrate that incorporating data diversity allows BLAST-based model training to achieve substantial advantages in both training efficiency and model performance.

In summary, the key contributions are as follows:
\begin{itemize}
    \item This study fills a critical gap in the role of data diversity in training universal forecasting models. It is the first to investigate the effect of pre-training data diversity on  training efficiency and model performance.
    \item 
    We propose a balanced sampling technique that treats time series patterns as the sampling target.
    Specifically, the time series is characterized by multiple statistical properties, and data is implicitly clustered using grid-based partitioning. Grid sampling and grid mixup techniques are then applied to generate diversified pre-training data.
    \item We develop BLAST, an efficient time series corpus generated through the balanced sampling. Experimental results show that BLAST-based pre-training achieves superior performance while reducing resource and data requirements.
\end{itemize}

%% file: sections/preliminaries.tex
\section{Preliminaries}
\label{sec_2}
In this section, we define the notions of large-scale time series forecasting datasets, sampling strategies, and universal forecasting models. Frequently used notations are summarized in Table \ref{tab:notations}.

\input{tables/Intro1}

\begin{definition}
\textbf{Large-scale Time Series Forecasting Dataset} $\mathcal{D}$ comprises $N$ sub-datasets, denoted as ${\mathcal{D}_1, \mathcal{D}_2, \dots, \mathcal{D}_N}$.
Each sub-dataset $\mathcal{D}_n$ contains $K_n$ time series $\{X_1^n, X_2^n, \dots, X_{K_n}^n\}$.
The $k$-th time series $X_k^n$ in the $n$-th sub-dataset consists of $T_{nk}$ time steps, denoted as $X_k^n = \{x_1^{nk}, x_2^{nk}, \dots, x_{T_{nk}}^{nk}\}$.
Note that the size of the sub-datasets and the length of individual time series can vary significantly.
\end{definition}

\begin{definition}
\textbf{Sampling Strategies} refer to the methods used to select training data from candidate sample set $\mathcal{W}$.
Raw time series cannot be directly used for model training.
Candidate samples $\mathcal{W}$ are generated by applying a sliding window $W$ to each time series.
The goal of the sampling strategy is to select the final set of samples used for training from these candidates.
\end{definition}

\begin{definition}
\textbf{Universal Forecasting Models}\footnote{While some studies refer to these models as foundational or general models, this paper adopts the term \textit{universal forecasting models}~\cite{MOIRAI, UniTS, GIFT_Eval, ROSE} for the sake of consistency and to avoid confusion with multi-task models.} are pre-trained on large-scale time series datasets and are capable of performing accurate zero-shot forecasting across diverse domains.
\end{definition}

%% file: tables/Intro1.tex
\begin{table}[t]
  \setlength{\abovecaptionskip}{0.2 cm}
    \caption{Comparison of training cost and performance between $\textbf{TimeMoE}_{base}$\textbf{-BLAST} and the original $\textbf{TimeMoE}_{base}$.
    The average MSE/MAE is reported as shown in Table~\ref{tab:main_result}.}
    \centering    
    \resizebox{0.95\linewidth}{!}{
        \begin{tabular}{c|c|c}
                \toprule[1.2pt]
                        & $\textbf{TimeMoE}_{base}\textbf{-BLAST}$ & $\textbf{TimeMoE}_{base}$  \\
                \midrule
        \textbf{Hardware}        & \textcolor{MyRed}{\underline{\textbf{8$\times$A100}}}        & \textcolor{black}{\textbf{128$\times$A100}} \\
                \midrule
        \textbf{\# Batch Size}      & \textcolor{MyRed}{\underline{\textbf{192}}}           & \textcolor{black}{\textbf{1024}}    \\
                \midrule
        \textbf{\# Training Tokens} & \textcolor{MyRed}{\underline{\textbf{78.64B}}}        &\textcolor{black}{\textbf{419.43B}}  \\
                \midrule
        \textbf{Avg. MSE / MAE}     &  \textcolor{MyRed}{\underline{\textbf{0.325 / 0.368}}} &  \textcolor{black}{{\textbf{0.341 / 0.385}}}   \\
                \bottomrule[1.2pt]
        \end{tabular}}
    \label{tab:Intro1}
\end{table}

%% file: sections/related_work.tex
\section{Related Work}

\subsection{Universal Time Series Forecasting}
Inspired by breakthroughs in artificial intelligence~\cite{GPT, LLaMA, TheInnovation1, TheInnovation2}, universal time series forecasting aim to achieve zero-shot forecasting across domains through pre-training on large-scale datasets.

These models are predominantly built on Transformer architectures~\cite{Transformer} and can be categorized into encoder-only models~\cite{MOIRAI, UniTS, MOMENT, ForecastPFN}, decoder-only models~\cite{LagLLaMA, Timer, TimesFM, TimeMoE, GPHT, TimeXL}, and encoder-decoder models~\cite{Chronos, TimeGPT1}. 
\textit{Encoder-only} models typically employ masked encoding strategies along with architectures tailored for time series tasks.
\textit{Decoder-only} models, on the other hand, often utilize autoregressive pre-training strategies.
Recent advancements have incorporated techniques such as mixture-of-experts~\cite{TimeMoE, MOIRAI-MoE}, long-context modeling~\cite{TimeXL}, and hierarchical modeling approaches~\cite{GPHT} to further improve their capabilities.
\textit{Encoder-decoder}~\cite{Chronos,TimeGPT1} architectures retain the full Transformer framework for time series tasks.
In parallel, cutting-edge research~\cite{TTM, DAM, ROSE} have begun exploring architectures beyond Transformers or other modalities~\cite{VisionTS,liu2025timekd,liu2025timecma}, aiming to design models specifically for time series data and further enhance forecasting accuracy.

Overall, these universal models demonstrate surprising zero-shot forecasting capabilities through pre-training on large-scale datasets,  underscoring their transformative potential in this field.

\subsection{Time Series Forecasting Pre-training Corpus}

Regardless of the model architectures, large-scale pre-training data $\mathcal{D}$ serves as the foundation for achieving universal forecasting. The size of the raw data is typically measured by the total number of observations, expressed as $\sum_{n=1}^N\sum_{k=1}^{K_n}T_{nk}$.

Numerous pioneering works have established large-scale training corpora to support universal forecasting models.
For instance, ForecastPFN~\cite{ForecastPFN} innovatively explored the role of purely synthetic data in pre-training.
Chronos~\cite{Chronos} combined data from sources such as Monash~\cite{Monash} and M-competitions~\cite{M5}, as well as synthetic data, to create a corpus with a total of 84 billion observations.
Similarly, MOIRAI~\cite{MOIRAI} introduced the large-scale dataset, LOTSA, which includes 231 billion observations (accounting for all variates).
Timer~\cite{Timer} developed the UTSD dataset by collecting multi-domain data, comprising 1 billion observations. 
Another example is TimeMoE~\cite{TimeMoE}, which constructed the largest existing dataset, Time-300B, by integrating various data sources, reaching a scale of 309 billion observations.
These contributions have laid a solid foundation for the development of universal forecasting models.

While most of these studies have focused primarily on the \textit{scale} of data, systematic investigations into \textit{diversity} remain unexplored.
To address this gap, we propose a diversified pre-training corpus—BLAST.
Built on 321 billion raw observations, BLAST leverages a balanced sampling strategy to ensure diversity.
We select state-of-the-art models and retrain them on the BLAST corpus.
Experimental results demonstrate that pre-training on BLAST is superior significantly in both training efficiency and model performance, underscoring the importance of a diversified corpus.  

\input{tables/preliminaries}

%% file: tables/preliminaries.tex
\begin{table}
\setlength{\abovecaptionskip}{0.2 cm}
\setlength{\belowcaptionskip}{-0.4cm}
\renewcommand\arraystretch{1.}
\caption{Frequently used notations.}
\label{tab:notations}
\resizebox{0.95\linewidth}{!}{
\begin{tabular}{m{1.3cm}<{\centering}|m{6.5cm}}
\toprule[1.2pt]
\textbf{Notations} & \textbf{Definitions}\\
\midrule
$\mathcal{D}$ & $\mathcal{D} = \{\mathcal{D}_1, \mathcal{D}_2, \dots, \mathcal{D}_N\}$ is the raw large-scale pre-training dataset, consisting of $N$ sub-datasets.\\
$\mathcal{D}_n$ & $\mathcal{D}_n = \{X_1^n, X_2^n, \dots, X_{K_n}^n\}$ is the $n$-th sub-dataset, containing $K_n$ time series. \\
$X_k^n$ & $X_k^n = \{x_1^{nk}, x_2^{nk}, \dots, x_{T_{nk}}^{nk}\}$ is the $k$-th time series in the $n$-th sub-dataset $\mathcal{D}_n$, containing $T_{nk}$ time steps.\\
$x_t^{nk}$ & $x_t^{nk}$ is the $t$-th time step in the time series $X_k^n$.\\
$\mathcal{W}$ & $\mathcal{W}$ denotes the collection of context windows drawn from $\mathcal{D}$.\\
$W$ & $W$ denotes the data under a context window of length $|W|$.\\
$S$ & $S$ denotes the stride of the sliding context window; throughout this paper we set $S = 1$ by default.\\
$\left\lfloor \cdot \right\rfloor$ & $\left\lfloor \cdot \right\rfloor$ denotes the floor operation.\\
\bottomrule[1.2pt]
\end{tabular}}
\end{table}

%% file: sections/method.tex
\begin{figure*}[t]
  \centering
    \setlength{\abovecaptionskip}{0. cm}
  \includegraphics[width=0.97\linewidth]{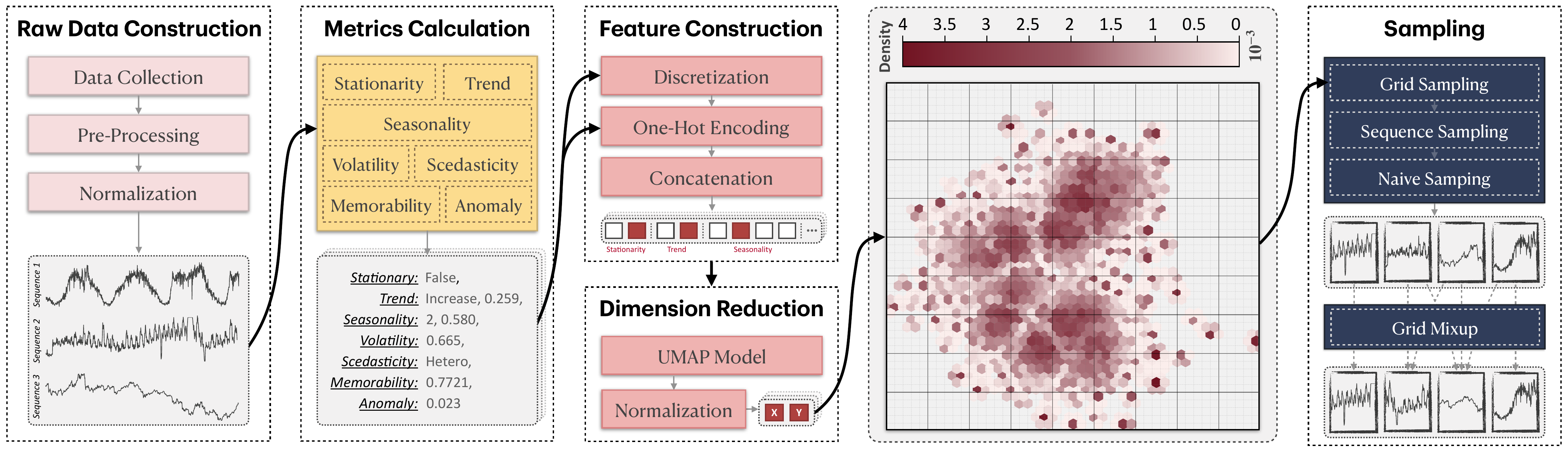}
  \caption{Pipeline for the balanced sampling: \underline{(i)} constructing large-scale time series datasets, \underline{(ii)} utilizing diverse metrics to comprehensively characterize time series, \underline{(iii)} generating unified feature vectors and performing dimension reduction to visualize data imbalances, and \underline{(iv)} implementing grid sampling and grid mixup to enhance the diversity of the training data.}
  \label{BLAST}
\end{figure*}

\section{Limitations of Existing Sampling Strategies}
\label{sec:4_1}

The purpose of a sampling strategy is to select training samples from the candidate sample set $\mathcal{W}$.
This set is generated by applying a sliding window with stride $S$ to each time series.
Each sample, denoted as $W_{n,k,t}$, corresponds the $t$-th sliding window position of the $k$-th time series in the $n$-th sub-dataset.
The sampling strategy is defined by the probability distribution $\mathbb{P}(W_{n,k,t})$.

\subsection{\textbf{Naive Sampling}}
\label{sec:naive_sampling}
The most straightforward way is the naive sampling, which uniformly selects the candidate samples:
\begin{equation}
    \mathbb{P}(W_{n,k,t})=\text{Uniform}(\mathcal{W})=\frac{1}{|\mathcal{W}|}.
\end{equation}
$\mathcal{W}$ is the candidate sample set, and is formally defined as:
\begin{equation}
    \begin{aligned}
        \mathcal{W} = \bigcup_{n=1}^N \bigcup_{k=1}^{K_n} \{W_{n,k,t} &\mid t = 1, 1+S, \dots, \text{and } t + |W| - 1 \leq T_{nk} \},\\
        |\mathcal{W}| &= \sum_{n=1}^N \sum_{k=1}^{K_n} \left\lfloor \frac{T_{nk} - |W|}{S} \right\rfloor + 1.
    \end{aligned}
\end{equation}
These notations are defined in Section \ref{sec_2} and Table~\ref{tab:notations}.

\input{tables/existing_corpus}

\subsection{\textbf{Stratified Sampling}}
\label{sec:strat_sampling}
Stratified sampling typically involves selecting a sub-dataset (uniformly or with weighted probabilities) and then applying naive sampling within it. The stratified sampling~(uniform) can be defined as:

\begin{equation}
    \begin{aligned}
    &\mathbb{P}(W_{n,k,t})=\mathbb{P}(\mathcal{W}_n)\cdot \mathbb{P}(W_{n,k,t}\mid \mathcal{W}_n), \\
    \mathbb{P}&(\mathcal{W}_n)=\frac{1}{N}, \quad\mathbb{P}(W_{n,k,t}\mid \mathcal{W}_n) = \frac{1}{|\mathcal{W}_n|}, 
    \end{aligned}
\end{equation}
where $N$ is the number of sub-datasets, $\mathcal{W}_n$ is the candidate sample set generate from sub-dataset $\mathcal{D}_n$, and can be defined as:
\begin{equation}
    \begin{aligned}
        \mathcal{W}_n = \bigcup_{k=1}^{K_n} \{W_{n,k,t} &\mid t = 1, 1+S, \dots, \text{and } t + |W| - 1 \leq T_{nk} \},\\
        &|\mathcal{W}_n| = \sum_{k=1}^{K_n} \left\lfloor \frac{T_{nk} - |W|}{S} \right\rfloor + 1.
    \end{aligned}
\end{equation}

\subsection{\textbf{The Limitations}}
An effective sampling strategy should generate samples with rich pattern while maintaining balanced sample sizes across patterns, \ie diversity.
However, naive sampling preserves the original data structure and its inherent biases.
Stratified sampling partially addresses this issue, but the assumption that domain or dataset labels reliably differentiate time series patterns is flawed.
Table~\ref{table:corpus_comparison} summarizes the corpus and sampling strategies in existing studies, most of which rely on naive sampling or stratified sampling, or their improved variants.
For instance, MOIRAI~\cite{MOIRAI} proposes the LOTSA dataset and employs a weighted stratified sampling approach with thresholds.
In summary, these simple strategies often lead to uneven data distributions, negatively affecting the model's convergence and generalization ability.

\section{Balanced Sampling Time Series Corpus}
\label{sec:4_2}
The core insight of BLAST lies in harnessing the diverse statistical characteristics of time series data to \textit{implicitly} cluster the data through grid-based partitioning.
Then, by treating the grids (\ie data patterns) as sampling units, BLAST employs grid sampling and grid mixup to sample the data in a balanced and comprehensive manner.
As illustrated in Figure \ref{BLAST}, BLAST involves several key processes: raw data construction, metrics calculation, feature construction, dimension reduction, and the sampling stage.

\subsection{\textbf{Raw Data Construction}}
We integrate extensive publicly available datasets, creating a large-scale dataset with a total of 321 billion observations.
We fill missing values with zeros and filter out short time series (those with a length of less than 512).
Commonly used benchmarks~\cite{Informer} are excluded.
Furthermore, we apply z-score normalization to eliminate the influence of varying value ranges across datasets.
See Appendix~\ref{appendix:A.1} for more details.

\subsection{\textbf{Metrics Calculation}}

As a core component of BLAST, metrics calculation serves to characterize a time series through a diverse set of metrics.
For a given time series $X$, BLAST utilizes seven statistical metrics, which characterize a time series' patterns from various aspects~\cite{TFB}.
Due to space limitations, additional details, including metrics selection principles, implementation details, and discussions on alternative methods, are provided in Appendix~\ref{appendix:A.2}.

\textit{Stationarity} refers to whether the statistical properties of a time series remain constant over time.
To assess this, we utilize the Augmented Dickey-Fuller (ADF) test, defined as:
\begin{equation}
    Stationary = \begin{cases} 
    \text{True}, & \text{if } \text{ADF}(X) < 0.05, \\ 
    \text{False}, & \text{otherwise}.
    \end{cases}
\end{equation}
The ADF test yields a boolean result, determining whether a given time series exhibits weak stationarity. Strong stationarity is not considered, as it is rarely encountered in real-world applications.

\textit{Trend}
describes the overall direction of change in a time series, reflecting long-term variation and representing a low-frequency component. To quantify the trend, we apply the Mann-Kendall test, formulated as:
\begin{equation}
    Trend, Strength_t = \text{MannKendall}(X).
\end{equation}
The \textit{Trend} can be classified as either \textit{increasing}, \textit{decreasing}, or \textit{no trend}, while \textit{Strength}$_t$ is a floating-point value that quantifies the magnitude or significance of the detected trend.

\textit{Seasonality}
represents recurrent fluctuations within a time series, characterized by high-frequency components.
We apply the Multiple Seasonal-Trend decomposition using Loess (M-STL)~\cite{MSTL} to decompose the time series into residual ($R$), trend ($T$), and multiple seasonal components ($S_i$):
\begin{equation}
    \begin{aligned}
    &[S_1, \cdots, S_{k}], T, R  = \text{M-STL}({X}),\\
    Str&ength_s = \text{max}(0, 1 - \frac{\text{var}(R)}{\text{var}(R + \sum_{1}^{k}S_{i})} ),
    \end{aligned}
\end{equation}
where \textit{Strength}$_s$ indicates the strength of seasonality.
We use the number of seasonal components, denoted as $k$ (with a maximum value of 3), along with \textit{Strength}$_s$, as the metrics.
Note that while the STL decomposition can also be used to calculate trends, doing so may result in redundancy between the trend and seasonality components, reducing their diversity.

\textit{Volatility} quantifies the degree of fluctuation in a time series and is formally defined as:
\begin{equation}
    Volatility = \frac{\sqrt{\frac{1}{T}\sum_{i=1}^{T}(x_i-\mu)^2}}{\mu},
\end{equation}
where $\mu$ is the mean of the time series with length $T$.
Essentially, volatility is a variation of the standard deviation, reflecting the relative magnitude of variability.

\textit{Scedasticity} indicates whether the variance of a time series changes over time, thereby capturing distribution drift.
It can be assessed using Lagrange Multiplier~(LM) test on the residual component~\cite{ARCH}:
\begin{equation}
    Scedasticity = \left\{
    \begin{aligned} 
    & \text{Homo},\quad \text{if\ \ LMTest}(R) > 0.05, \\
    & \text{Hetero},\quad \text{otherwise}.\\
    \end{aligned} 
    \right.
\end{equation}

\textit{Memorability} quantifies the degree of long-term dependence in a time series and is measured using the Hurst exponent:
\begin{equation}
	Memorability = Hurst(X).
\end{equation}

\textit{Anomaly} 
represents the proportion of values that deviate significantly from the majority, reflecting the level of noise in the series.
Outliers are identified as values exceeding the 95\% threshold in a one-tailed test after z-score normalization:
\begin{equation}
    Anomaly = \frac{|\{x_{i} \in X | \frac{x_{i} - \mu}{\sigma} > 1.645\}|}{T}.
\end{equation}

\subsection{\textbf{Feature Construction}}
\label{4.2.3}

\begin{table}[b]
\setlength{\belowcaptionskip}{-0.cm}
\setlength{\abovecaptionskip}{0.2 cm}
\setlength{\tabcolsep}{1.5pt}
\caption{Discretization of continuous metrics.}
\resizebox{\linewidth}{!}{
\begin{tabular}{c|c|c|c|c|c}
\toprule[1.2pt]
\textbf{Metric} & \textbf{\textit{Strength$_t$}} & \textbf{\textit{Strength$_s$}} & \textbf{\textit{Volatility}} & \textbf{\textit{Memorability}} & \textbf{\textit{Anomaly}} \\
\midrule
B      & 20          & 10          & 6          & 10           & 4       \\
\midrule
$b_0$  & -1          & 0           & 0          & 0            & 0       \\
\midrule
$b_B$  & 1           & 1           & 1.2        & 1            & 0.16   \\
\bottomrule[1.2pt]
\end{tabular}
}
\label{table:discrete}
\end{table}

\begin{figure}[t]
  \centering
  \setlength{\abovecaptionskip}{0.2cm}
  \setlength{\belowcaptionskip}{-0.4cm}
  \includegraphics[width=0.95\linewidth]{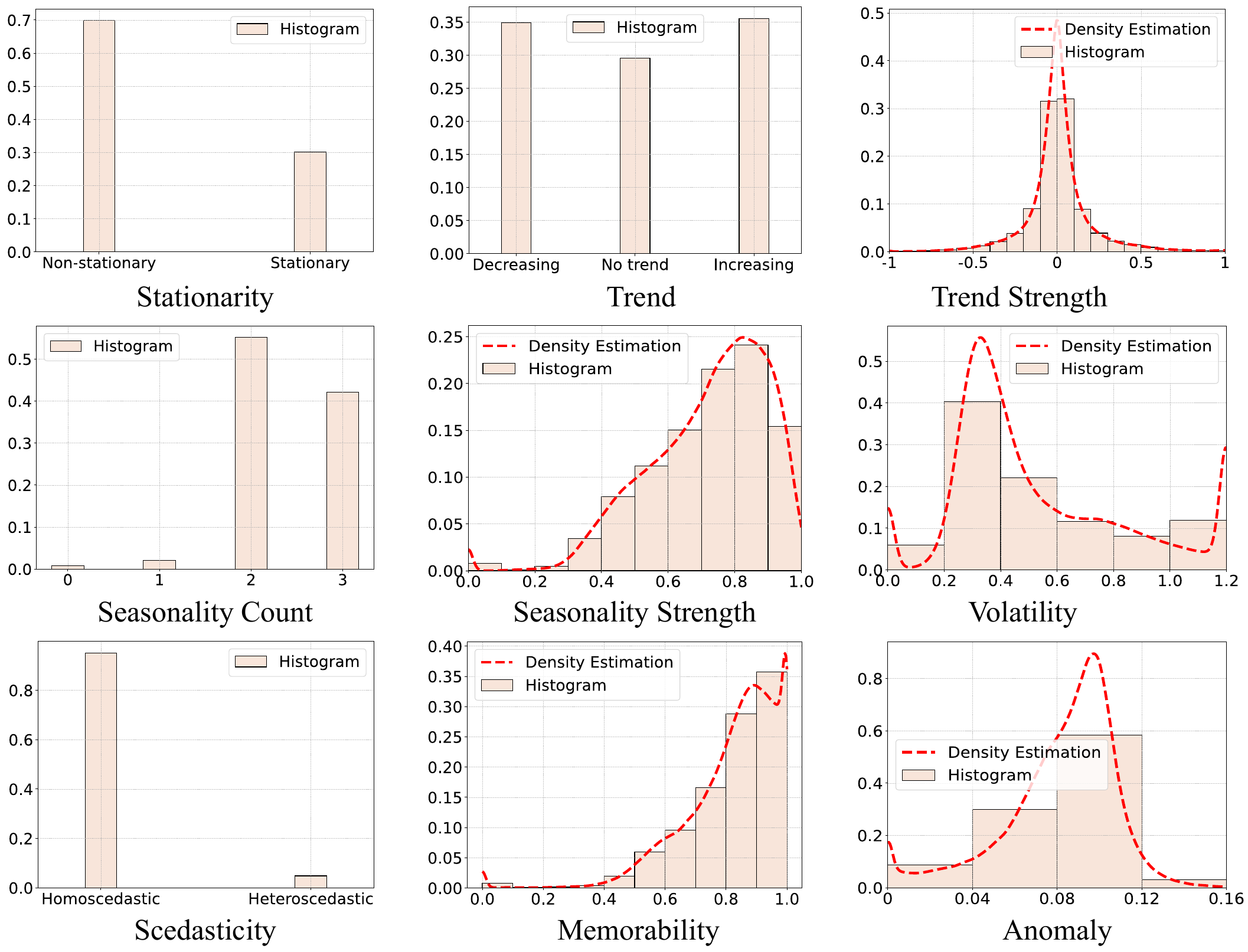}
  \caption{Distribution of the raw dataset across key metrics.}
  \label{fig:distribution}
\end{figure}

Overall, the metrics described above provide a comprehensive characterization of a time series.
Figure~\ref{fig:distribution} illustrates the distribution of the raw data across these metrics.
As can be seen, these metrics are inherently heterogeneous, comprising both discrete and floating-point values with varying ranges.
To mitigate this heterogeneity, we introduce a discretization-based feature construction approach that unifies the representation of all metrics into a single vector.  

For continuous metrics, we discretize their values within a predefined range using a quantization technique.
Formally, inspired by~\cite{Chronos}, given a metric $z$, the interval $[b_0, b_B]$ is divided into $B$ equally spaced bins, and $z$ is mapped to the corresponding bin index using the quantization function $g(z)$, defined as follows:  
\begin{equation}
    g(z) = 
    \begin{cases} 
    0, & \text{if } b_0 \leq z < b_1, \\
    1, & \text{if } b_1 \leq z < b_2, \\
    \vdots & \\
    B-1, & \text{if } b_{B-1} \leq z < b_B. \\
    \end{cases}
\end{equation}
Values outside the interval $[b_0, b_B]$ are assigned to the nearest bin (either 0 or $B-1$) to handle the long-tail distribution.
The parameters $B$, $b_0$, and $b_B$ for each continuous metric are listed in Table~\ref{table:discrete}.

Finally, along with discrete metrics, we apply one-hot encoding to all metrics.
These vectors are then concatenated into a unified representation $h$, which has a fixed length of 61, \ie a vector in $\mathbb{R}^{61}$, providing a standardized and comprehensive description of the time series patterns.

\subsection{\textbf{Dimension Reduction}}
\label{4.2.4}
To better understand the bias in the data distribution, we reduce the dimension of the vector $h$ to a low-dimensional space.
Specifically, for a given time series $X_k^n$, its corresponding vector $h$ can be calculated.
The BLAST raw dataset comprises approximately 40 million raw time series.
Subsequently, we employ the UMAP~\cite{UMap} model $f_{\text{umap}}$ to project all sparse vectors $h$ into a dense two-dimensional space. Compared with other dimension reduction techniques such as t-SNE~\cite{t-SNE} and PCA~\cite{PCA}, UMAP offers the advantages of higher efficiency and better preservation of data structure. The transformation is expressed as follows:  
\begin{equation}
h' = f_{\text{umap}}(h) \in \mathbb{R}^2,
\end{equation}
where $h$ represents the original vector, and $h'$ is the corresponding vector after dimension reduction. We normalize all $h'$ to $[0, 1]$.
Due to space constraints, the details of the UMAP model's implementation and hyper-parameter study are provided in Appendix~\ref{appendix:A.3}.

As shown in Figure~\ref{BLAST}, the reduced data reveals a clear global structural pattern, though its distribution remains highly imbalanced. This skewed distribution can introduce bias during model training, as discussed in Section~\ref{sec_1}. Furthermore, the gaps between different regions suggest that the patterns in the raw dataset are still insufficient, despite the large scale of the data.

\subsection{\textbf{Sampling}}
To address the issue of uneven data distribution, we propose an intuitive and effective sampling approach, which incorporates both grid sampling and grid mixup.

First, we uniformly partition the two-dimensional space ($x, y \in [0, 1]$) into $M \times M$ grids, denoted as $\mathcal{G}$, with each grid containing multiple time series.
Grid sampling is then applied, which involves first selecting a grid, then randomly sampling a time series within that grid, followed by naive sampling. The probability of selecting a sample $W_{n,k,t}$ is given by the following:
\begin{equation}
    \begin{aligned}
    \mathbb{P}(W_{n,k,t}) = \mathbb{P}(\mathcal{G}_m)\cdot &\mathbb{P}(\mathcal{W}_{n, k}\mid \mathcal{G}_m)\cdot \mathbb{P}(W_{n,k,t}\mid \mathcal{W}_{n, k}),\\
    &\mathbb{P}(\mathcal{G}_m)=\frac{1}{|\mathcal{G}|},\\
    \mathbb{P}(\mathcal{W}_{n, k}\mid\mathcal{G}_m)=\frac{1}{|\mathcal{G}_m|}, &\  \mathbb{P}(W_{n,k,t}\mid \mathcal{W}_{n, k})=\frac{1}{|\mathcal{W}_{n, k}|}, \forall X_k^n\in\mathcal{G}_m,\\
    \end{aligned}
\end{equation}
where $|\mathcal{G}|$ is the number of valid grids 
, $|\mathcal{G}_m|$ represents the number of time series included in $m$-th grid, and $\mathcal{W}_{n,k}$ is the candidate sample set generate from time series $X_k^n$. We set $M=100$.

Next, to address the lack of sufficient coverage, \ie the gaps between different regions of the data distribution, we introduce a grid mixup technique that further enhances the model's generalization ability.
Specifically, we randomly pick $k$ grids (from all available grids), where $k$ is drawn from the discrete uniform distribution $\mathcal{U}(1, K)$, and then randomly select samples from these grids.
These samples are subsequently mixed as follows:
\begin{equation}
    {X}^{\text{\scriptsize GridMixup}} = \sum_{i=1}^{k}\lambda_i X^i,
\end{equation}
where $X^i$ is the sample from grid $i$, and $[\lambda_1, \cdots, \lambda_k]$ are sampled from a symmetric Dirichlet distribution $D(\alpha)$, where $\alpha=1.5$.
We set $K=3$, \ie the original data remains in the dataset with a 33.33\% probability.
This approach is inspired by TSMixup~\cite{Chronos}, but instead of treating each time series as the basic unit of sampling, we use the grid as the fundamental sampling unit.

In summary, the sampling stage mitigates bias in over-dense or under-dense regions, effectively addressing biases in large-scale datasets. This strategy ensures that the samples are balanced and representative, thereby enhancing both the efficiency and generalization performance of the model training process.

%% file: tables/existing_corpus.tex
\begin{table}[b]
\setlength{\abovecaptionskip}{0.2 cm}
\renewcommand\arraystretch{1.}
\caption{Comparison of time series corpora.}
\resizebox{\linewidth}{!}{
    \begin{tabular}{l|c|c|c}
    \toprule[1.2pt]
    \textbf{Corpus}            & \textbf{Raw Size} & \textbf{Open Source} & \textbf{Sampling Strategy}         \\
    \midrule
    UTSD~\cite{Timer}          & 1B                               & \checkmark            & Naive Sampling                    \\
    MOMENT~\cite{MOMENT}       & 1.23B                            & \checkmark            & Naive Sampling                    \\
    Chronos~\cite{Chronos}     & 84B                              & \checkmark            & Stratified Sampling               \\
    TimeGPT~\cite{TimeGPT1}    & $\sim$100B                       & $\times$              & Unknown                            \\
    LOTSA~\cite{MOIRAI}        & 231B                             & \checkmark            & Stratified Sampling               \\
    TimesFM~\cite{TimesFM}      & $\sim$307B                       & $\times$              & Unknown                            \\
    Time-300B~\cite{TimeMoE}   & 309B                             & \checkmark            & Naive Sampling                    \\
    \midrule
    \textbf{BLAST}                      & \textbf{321B}                             & \textbf{\checkmark}            & \textbf{Balanced Sampling}                 \\
    \bottomrule[1.2pt]
    \end{tabular}}
\label{table:corpus_comparison}
\end{table}

%% file: sections/experiments.tex
\section{Experiments}
This section addresses the following key research questions through comprehensive experiments:

\begin{itemize}[leftmargin=*]
    \item \textbf{RQ1:} Does pre-training on BLAST provide any advantages?
    \item \textbf{RQ2:} What are the sources of these advantages, and what is the impact of different sampling strategies?
    \item \textbf{RQ3:} How do grid sampling and grid mixup influence balanced sampling (through ablation and hyperparameter analysis)?
\end{itemize}

\input{tables/main_results}
\input{tables/gift_eval}

\subsection{Experimental Setup}
\subsubsection{\textbf{Baselines}}
\label{subsection:baselines}
We select three popular universal forecasting models—TimeMoE~\cite{TimeMoE}, MOIRAI~\cite{MOIRAI}, and Chronos~\cite{Chronos}. For TimeMoE and MOIRAI, we consider both their base and large versions, whereas for Chronos\footnote{We employ the latest Chronos‑Bolt release for its superior efficiency and accuracy.} we include the small and base versions. This yields six baselines in total.
The dataset sizes and sampling methods originally used in their respective paper are detailed in Table~\ref{table:corpus_comparison}.

\subsubsection{\textbf{Datasets}}
\label{subsection:datasets}
Following TimeMoE~\cite{TimeMoE}, we select six commonly used benchmarks: ETTh1, ETTh2, ETTm1, ETTm2, Weather, and GlobalTemp. None of these datasets is included in BLAST.
Additionally, we adopt GIFT-Eval~\cite{GIFT_Eval}, the latest comprehensive benchmark containing 97 small prediction tasks.
After filtering out any data present in Time‑300B (TimeMoE pre‑training data), LOTSA (MOIRAI pre‑training data), and BLAST, we use the remaining 43 tasks based on the original GIFT-Eval settings.

\subsubsection{\textbf{Implementation Details}}
All experiments are conducted using PyTorch on $8\times$A100 GPUs (40GB).
The code for training universal forecasting models with BLAST is available at {\textit{\color{blue}\url{https://github.com/GestaltCogTeam/BasicTS}}}, and the BLAST corpus generation code can be found at \textit{\color{blue}{\url{https://github.com/GestaltCogTeam/BLAST}}}.
Additionally, all subsequent experimental results follow the {\textit{zero-shot}} forecasting setting.
Further implementation details for the benchmark datasets are provided in Appendix~\ref{appendix:B}.

\input{tables/ablation}

\subsection{Pre-training on BLAST (RQ1)}
\label{sec:rq1}
This section evaluates the advantages of training universal forecasting models using the BLAST corpus. To achieve this, we retrain each of the selected baselines, as detailed in \S \ref{subsection:baselines}, \textbf{from scratch} using the BLAST corpus. We then compare the performance of the retrained models with their pre-trained counterparts on the benchmarks outlined in \S \ref{subsection:datasets}.

\textit{\textbf{Settings.}}
We adhered to the original setup as outlined in their respective papers~\cite{Chronos, TimeMoE, MOIRAI}. 
Due to space limitations, readers interested in more details can refer to the original papers.
The only deviation from the original setup was the batch size for TimeMoE. The original TimeMoE model~\cite{TimeMoE} was trained using \textbf{128$\times$A100 GPUs} with a \textbf{batch size of 1024}, processing \textbf{419.43 billion training tokens}. Thanks to its massive model parameters and training data, it achieved state-of-the-art performance.
However, due to computational resource constraints, the TimeMoE model pre-trained on BLAST used a reduced \textbf{batch size of 192}, training on \textbf{78.64 billion tokens}.
For the benchmarks used in TimeMoE~\cite{TimeMoE}, we follow a similar setup. We assess the performance across four different prediction lengths: $[96, 192, 336, 720]$. We report the normalized Mean Squared Error~(MSE) and Mean Absolute Error~(MAE).
For the GIFT-Eval benchmark~\cite{GIFT_Eval}, we filtered out data already included in Time-300B (TimeMoE pre-training data), LOTSA (MOIRAI pre-training data), and BLAST, and strictly followed its evaluation pipeline. We report the Mean Absolute Scaled Error~(MASE).

\textit{\textbf{Results.}}
Table~\ref{tab:main_result} and Table~\ref{tab:gift_eval} present the results of our experiments.
In general, models pre-trained on the BLAST corpus outperform the original models.
The results for TimeMoE highlight the significant efficiency advantages brought by pre-training on BLAST, both in terms of computational resources and data usage. Specifically, BLAST-based pre-training requires only 8 A100 GPUs, compared to 128 A100 GPUs for the original TimeMoE, and processes 78.64 billion training tokens, which is a fraction of the 419.43 billion tokens required for the original model.
Furthermore, the results for MOIRAI and Chronos demonstrate that, when computational resources and the number of training tokens are similar, the performance advantages brought by BLAST become even more apparent.
In the next part, we delve deeper into the impact of sampling strategies on both training efficiency and model performance.  

\subsection{Impact of Sampling Strategies (RQ2)}
This section provides a comprehensive analysis of the impact of different sampling strategies on \textbf{{training efficiency}} and \textbf{{predictive performance}}, shedding light on the key factors contributing to the advantages of BLAST pre-training. To quantify the effects of each sampling strategy precisely, we conduct controlled experiments \textbf{using the same raw data.}

\textbf{\textit{Settings.}} We use $\text{TimeMoE}_{base}$ as the baseline model, with experimental configurations consistent with \S \ref{sec:rq1}. Based on the raw BLAST data, $\text{TimeMoE}_{base}$ are trained using datasets derived from three different sampling strategies: naive sampling~(\S \ref{sec:naive_sampling}), stratified sampling~(\S \ref{sec:strat_sampling}), and balanced sampling~(\S \ref{sec:4_2}).
First, to evaluate the effects of these strategies on training efficiency, we analyze the rate of validation loss reduction on a unified validation set, which is constructed as the union of the validation sets from the three sampling strategies and excludes data that appears in the training sets.
Second, to assess the impact of sampling strategies on model performance, we report the MAE on four ETT datasets, enabling a comprehensive comparison across different sampling methods.

\begin{figure}[t]
  \centering
  \setlength{\abovecaptionskip}{0.2cm}
  \setlength{\belowcaptionskip}{-0.2cm}
  \includegraphics[width=0.9\linewidth]{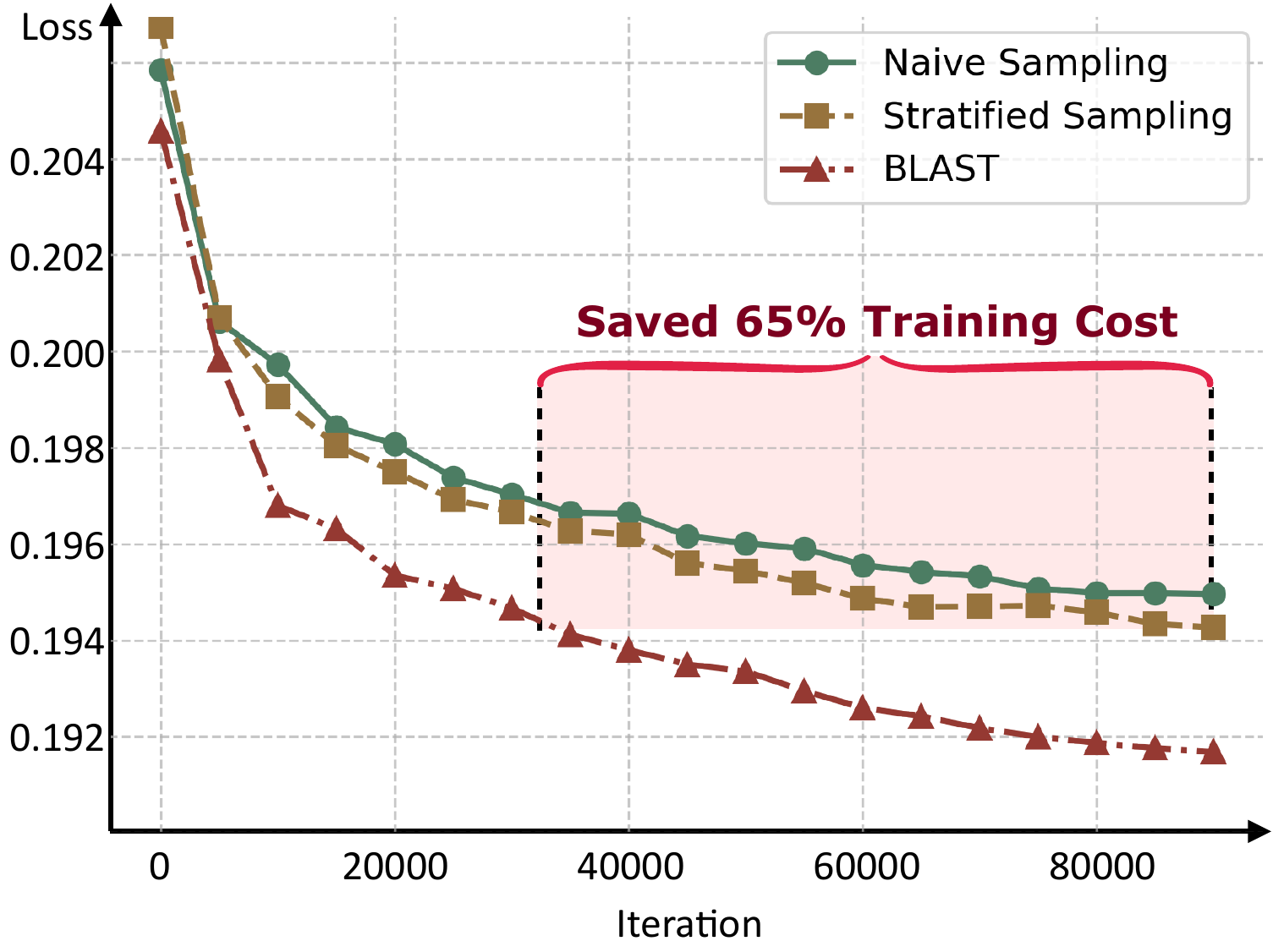}
  \caption{Comparison of convergence speeds for different sampling methods.}
  \label{ablation:fig}
\end{figure}

\textbf{\textit{Results.}}
Figure~\ref{ablation:fig} illustrates the convergence rates of models trained with different sampling strategies, highlighting their effects on training efficiency. The results indicate that models trained with balanced sampling exhibit a significantly faster reduction in loss. This efficiency advantage becomes particularly evident in the later stages of training, where loss reduction slows. Notably, under equivalent loss conditions, balanced sampling requires only about 35\% of the training steps compared to naive or stratified sampling.
Table~\ref{ablation:tab} further demonstrates the effectiveness of different sampling methods in terms of forecasting performance.
Balanced sampling consistently outperforms other methods.
Additionally, models trained with naive or stratified sampling underperform the original TimeMoE. This is due to the lack of focus on data diversity in these strategies, combined with the substantially smaller token count in our training process compared to the original TimeMoE implementation.

In summary, these results underscore the critical role of data diversity, and the balanced sampling strategy significantly enhances data diversity during training.
This intuitive yet effective sampling approach proves instrumental in improving both model performance and training efficiency.

\subsection{How Do Grid Sampling and Grid Mixup Affect Balanced Sampling? (RQ3)}
This part presents a further ablation study and hyper-parameter analysis of BLAST.
Specifically, we examine the contributions of two key components—grid sampling and grid mixup—in balanced sampling.
Additionally, we investigate how grid size affects model performance and explore the underlying reasons for these effects.

\textbf{\textit{Settings.}}
We use TimeMoE$_{base}$ as the baseline model and conduct experiments on datasets excluding either grid sampling or grid mixup.
We report the MAE on four ETT datasets.
Furthermore, we vary the grid size in the sampling stage, setting it to $[10, 50, 100, 500, 1000 , 5000]$.
We evaluate the models on four ETT datasets and report their averaged predictive performance.

\begin{figure}[t]
  \centering
  \setlength{\abovecaptionskip}{0.2cm}
  \includegraphics[width=1\linewidth]{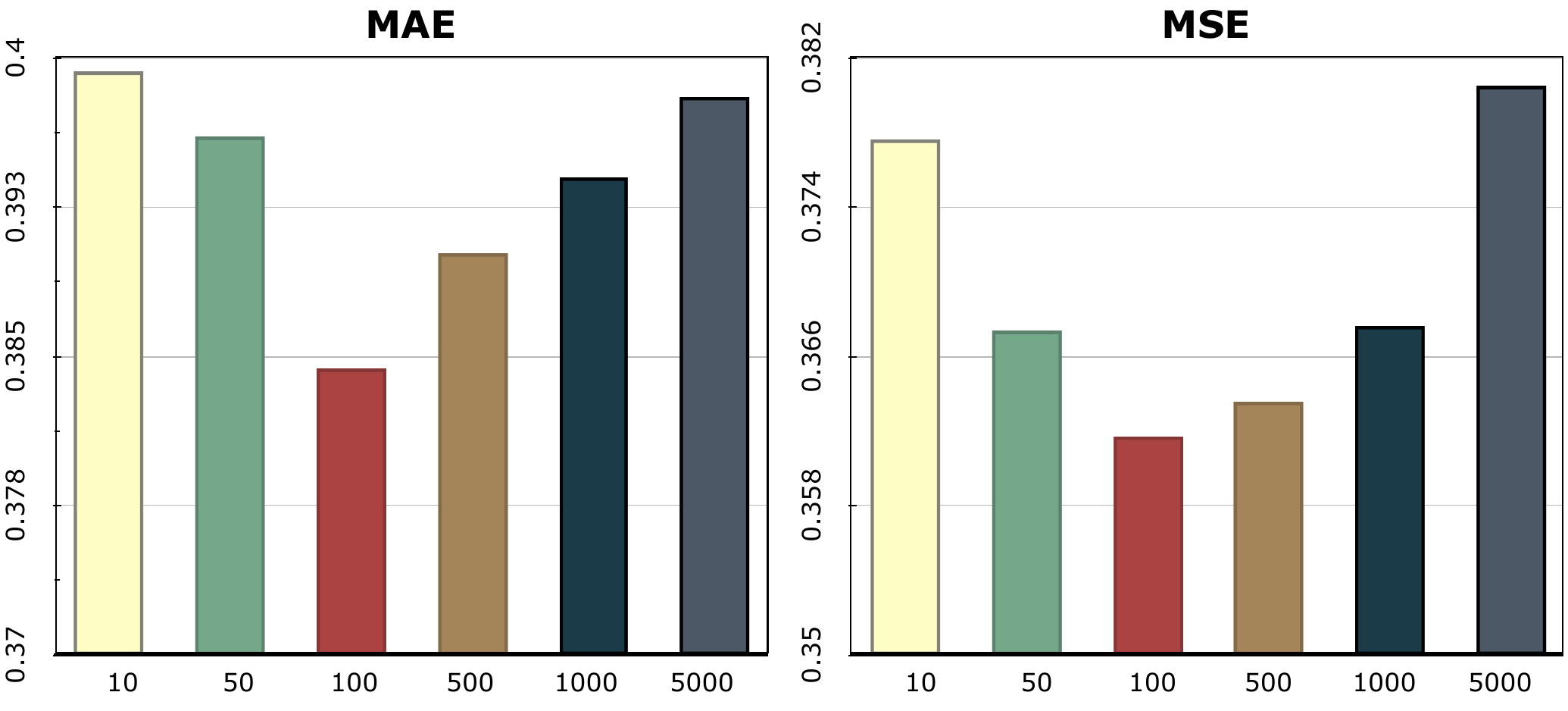}
  \caption{The impact of grid size in grid sampling.}
  \label{grid_size}
\end{figure}

\textbf{\textit{Results.}}
The performance of models without grid sampling and grid mixup is shown in Table~\ref{ablation:tab}.
Removing grid sampling results in a setup similar to naive sampling, where grid mixup becomes a standard TSMixup~\cite{Chronos}.
This yields slightly better performance than naive sampling.
Meanwhile, the absence of grid mixup significantly diminishes performance compared to balanced sampling.
This confirms that grid mixup is an effective strategy for enhancing data diversity.
These ablation results further validate the effectiveness of balanced sampling.

Additionally, Figure 6 presents the predictive performance for various grid sizes.
It is evident that both excessively small and large grids result in suboptimal performance. 
The reaseaons are:

\begin{itemize}[leftmargin=10pt]
    \item Too large a grid: Results in too few grids, each with many heterogeneous time series. In the extreme case, there's just one grid, and balanced sampling degrades to naive sequence sampling.
    \item Too small a grid: Results in too many grids. Despite large data, the representation space remains sparse, and balanced sampling degrades to naive sequence sampling again due to insufficient sequences per grid.
\end{itemize}
In summary, grid size acts like implicit clustering—ineffective clustering (either too large or too small grid size) causes balanced sampling to fail.

\subsection{Alternative Dimension Reduction Methods}
\label{sec:6_5}
We benchmark three popular dimensionality-reduction algorithms, PCA \cite{PCA}, t-SNE \cite{t-SNE}, and UMAP.
To obtain an intuitive sanity check, we generated synthetic data with uniformly distributed feature vectors, following the feature construction process in BLAST.
If a method faithfully preserves the original geometry, its projection should therefore exhibit:
\begin{itemize}[leftmargin=10pt]
    \item Clear global structure, as the unified vector is constructed from multiple one-hot vectors.
    \item Even distribution of samples within each component, as each one-how vector is randomly generated.
\end{itemize}
To compare these methods, we visualized the results for both real and synthetic data. 

Table \ref{pca_tsne_umap} contrasts the three dimensionality-reduction methods.
Because PCA is a linear method, it fails to represent either local or global structure in our discrete feature vectors.
t-SNE preserves neighbourhoods but distorts the overall geometry and, on large datasets, is computationally heavy.
UMAP, by comparison, captures both global relationships and local patterns: it separates the main regions cleanly and keeps the samples within each region evenly distributed. Overall, UMAP provides the most faithful picture of the data at both macro- and micro-scales.

\begin{table}[t]
  \centering
    \setlength{\abovecaptionskip}{0.2 cm}
  \caption{Comparison between PCA, t-SNE, and UMAP.}
  \includegraphics[width=1\linewidth]{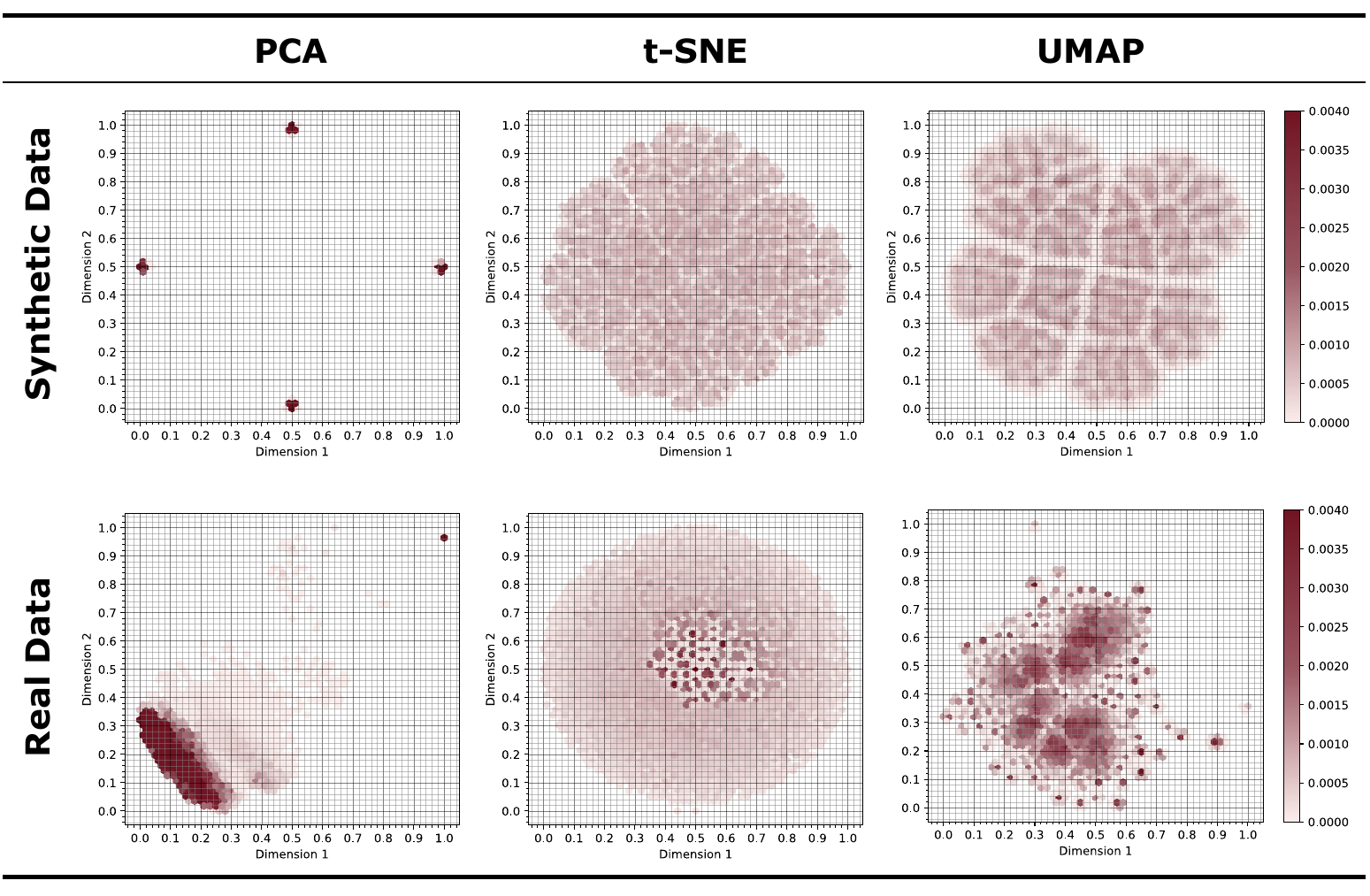}
  \label{pca_tsne_umap}
\end{table}

\subsection{Intuition Behind Balanced Sampling}

Essentially, BLAST estimates the probability density function (PDF) of the pre-training data and then draws unbiased samples via stratified sampling guided by that PDF.
Directly estimating a PDF for raw time series data is impractical because each series is a high-dimensional vector.
BLAST therefore compresses every series into a small set of statistical descriptors and projects these descriptors into a two-dimensional feature space.
It then partitions this plane into uniform grid cells and samples within them.
Each cell implicitly defines a cluster—capturing a characteristic pattern—so the cells themselves, rather than pre-defined classes or domain labels, become the strata for sampling.

Additionally, explicit clustering algorithms such as k-means or DBSCAN could serve the same purpose, but they scale poorly and often fail to preserve cluster quality on large datasets. Grid sampling offers a more intuitive, computationally lightweight alternative that strikes a practical balance between simplicity and effectiveness.

%% file: tables/main_results.tex
\begin{table*}[]
\setlength{\abovecaptionskip}{.2 cm}
\caption{
Performance comparison of BLAST retrained models with their pretrained counterparts.
Lower MAE and MSE values indicate superior performance.
The symbols $s$, $b$, and $l$ represent the small, base, and large versions, respectively.
$\dagger$ denotes the models retrained from scratch using the BLAST corpus. Models with superior or equal performance are highlighted in {\color{MyRed2}\textbf{red}}.
}
\setlength\tabcolsep{1.2pt}
\resizebox{1\linewidth}{!}{
\begin{tabular}{rr|cccc|cccc|cccc|cccc|cccc|cccc}
\toprule[1.5pt]
\multicolumn{2}{l}{\textbf{Models}} & \multicolumn{2}{|l}{\textbf{TimeMoE}$_{l}^\dagger$} & \multicolumn{2}{l|}{\textbf{TimeMoE}$_{l}$} & \multicolumn{2}{l}{\textbf{TimeMoE}$_{b}^\dagger$} & \multicolumn{2}{l|}{\textbf{TimeMoE}$_{b}$} & \multicolumn{2}{l}{\textbf{MOIRAI}$_{l}^\dagger$} & \multicolumn{2}{l|}{\textbf{MOIRAI}$_{l}$} & \multicolumn{2}{l}{\textbf{MOIRAI}$_{b}^\dagger$} & \multicolumn{2}{l|}{\textbf{MOIRAI}$_{b}$} & \multicolumn{2}{l}{\textbf{Chronos}$_{b}^\dagger$} & \multicolumn{2}{l|}{\textbf{Chronos}$_{b}$} & \multicolumn{2}{l}{\textbf{Chronos}$_{s}^{\dagger}$} & \multicolumn{2}{l}{\textbf{Chronos}$_{s}$} \\
\cmidrule(r){3-4} \cmidrule(r){5-6} \cmidrule(r){7-8} \cmidrule(r){9-10} \cmidrule(r){11-12} \cmidrule(r){13-14} \cmidrule(r){15-16} \cmidrule(r){17-18} \cmidrule(r){19-20} \cmidrule(r){21-22} \cmidrule(r){23-24} \cmidrule(r){25-26}
\multicolumn{2}{l|}{{Metrics}} & MSE & MAE & MSE & MAE & MSE & MAE & MSE & MAE & MSE & MAE & MSE & MAE & MSE & MAE & MSE & MAE & MSE & MAE & MSE & MAE & MSE & MAE & MSE & MAE \\
\midrule
\multirow{5}{*}{\rotatebox{90}{ETTh1}} & 96 &  {\color{MyRed2}\textbf{.348}} &  {\color{MyRed2}\textbf{.375}} & .350 & .382 &  {\color{MyRed2}\textbf{.352}} &  {\color{MyRed2}\textbf{.376}} & .357 & .381 &  {\color{MyRed2}\textbf{.359}} &  {\color{MyRed2}\textbf{.383}} & .381 & .388 &  {\color{MyRed2}\textbf{.362}} &  {\color{MyRed2}\textbf{.384}} & .376 & .392 &  {\color{MyRed2}\textbf{.357}} &  {\color{MyRed2}\textbf{.375}} & .384 & .379 &  {\color{MyRed2}\textbf{.359}} & {\color{MyRed2}\textbf{.376}} & .394 &  .381 \\
 & 192 &  {\color{MyRed2}\textbf{.381}} &  {\color{MyRed2}\textbf{.399}} & .388 & .412 & .389 &  {\color{MyRed2}\textbf{.401}} &  {\color{MyRed2}\textbf{.384}} & .404 &  {\color{MyRed2}\textbf{.395}} &  {\color{MyRed2}\textbf{.404}} & .434 & .415 &  {\color{MyRed2}\textbf{.400}} &  {\color{MyRed2}\textbf{.405}} & .412 & .413 &  {\color{MyRed2}\textbf{.397}} &  {\color{MyRed2}\textbf{.401}} & .441 & .412 &  {\color{MyRed2}\textbf{.403}} &  {\color{MyRed2}\textbf{.402}} & .455 &  .414 \\
 & 336 &  {\color{MyRed2}\textbf{.409}} &  {\color{MyRed2}\textbf{.424}} & .411 & .430 &  {\color{MyRed2}\textbf{.408}} &  {\color{MyRed2}\textbf{.419}} & .411 & .434 &  {\color{MyRed2}\textbf{.411}} &  {\color{MyRed2}\textbf{.416}} & .495 & .445 &  {\color{MyRed2}\textbf{.416}} &  {\color{MyRed2}\textbf{.420}} & .433 & .428 &  {\color{MyRed2}\textbf{.422}} &  {\color{MyRed2}\textbf{.417}} & .475 & .430 &  {\color{MyRed2}\textbf{.431}} &  {\color{MyRed2}\textbf{.416}} & .499 & .444 \\
 & 720 & .447 &  {\color{MyRed2}\textbf{.451}} &  {\color{MyRed2}\textbf{.427}} & .455 & .450 &  {\color{MyRed2}\textbf{.455}} &  {\color{MyRed2}\textbf{.449}} & .477 &  {\color{MyRed2}\textbf{.420}} &  {\color{MyRed2}\textbf{.430}} & .611 & .510 &  {\color{MyRed2}\textbf{.430}} &  {\color{MyRed2}\textbf{.439}} & .447 & .444 &  {\color{MyRed2}\textbf{.460}} &  {\color{MyRed2}\textbf{.443}} & .472 & .446 &  {\color{MyRed2}\textbf{.449}} &  {\color{MyRed2}\textbf{.439}} & .520 & .476 \\
 \rowcolor{tabhighlight}\cellcolor{white}& AVG & .396 &  {\color{MyRed2}\textbf{.412}} &  {\color{MyRed2}\textbf{.394}} & .419 &  {\color{MyRed2}\textbf{.399}} &  {\color{MyRed2}\textbf{.412}} & .400 & .424 &  {\color{MyRed2}\textbf{.396}} &  {\color{MyRed2}\textbf{.408}} & .480 & .439 &  {\color{MyRed2}\textbf{.402}} &  {\color{MyRed2}\textbf{.412}} & .417 & .419 &  {\color{MyRed2}\textbf{.409}} & {\color{MyRed2}\textbf{.409}} & .443 & .416 &  {\color{MyRed2}\textbf{.410}} &  {\color{MyRed2}\textbf{.408}} & .467 & .428 \\
\midrule
\multirow{5}{*}{\rotatebox{90}{ETTh2}} & 96 & {\color{MyRed2}\textbf{.276}} & {\color{MyRed2}\textbf{.329}} & .302 & .354 & {\color{MyRed2}\textbf{.285}} & {\color{MyRed2}\textbf{.332}} & .305 & .359 & {\color{MyRed2}\textbf{.288}} & {\color{MyRed2}\textbf{.325}} & .296 & .330 & {\color{MyRed2}\textbf{.284}} & {\color{MyRed2}\textbf{.324}} & .294 & .325 & {\color{MyRed2}\textbf{.282}} & {\color{MyRed2}\textbf{.321}} & .289 & .330 & {\color{MyRed2}\textbf{.281}} & {\color{MyRed2}\textbf{.326}} & .282 & .328 \\
 & 192 & {\color{MyRed2}\textbf{.345}} & {\color{MyRed2}\textbf{.376}} & .364 & .385 & {\color{MyRed2}\textbf{.348}} & {\color{MyRed2}\textbf{.378}} & .351 & .386 & {\color{MyRed2}\textbf{.353}} & {\color{MyRed2}\textbf{.370}} & .361 & .371 & {\color{MyRed2}\textbf{.348}} & {\color{MyRed2}\textbf{.369}} & .365 & .375 & {\color{MyRed2}\textbf{.356}} & {\color{MyRed2}\textbf{.369}} & .359 & {\color{MyRed2}\textbf{.369}} & {\color{MyRed2}\textbf{.353}} & {\color{MyRed2}\textbf{.371}} & .354 & .373 \\
 & 336 & {\color{MyRed2}\textbf{.384}} & {\color{MyRed2}\textbf{.416}} & .417 & .425 & {\color{MyRed2}\textbf{.372}} & {\color{MyRed2}\textbf{.405}} & .391 & .418 & {\color{MyRed2}\textbf{.369}} & {\color{MyRed2}\textbf{.382}} & .390 & .390 & {\color{MyRed2}\textbf{.367}} & {\color{MyRed2}\textbf{.386}} & .376 & .390 & {\color{MyRed2}\textbf{.378}} & {\color{MyRed2}\textbf{.397}} & .399 & .400 & {\color{MyRed2}\textbf{.387}} & {\color{MyRed2}\textbf{.403}} & .416 & .410 \\
 & 720 & {\color{MyRed2}\textbf{.442}} & {\color{MyRed2}\textbf{.470}} & .537 & .496 & {\color{MyRed2}\textbf{.419}} & {\color{MyRed2}\textbf{.452}} & {\color{MyRed2}\textbf{.419}} & .454 & {\color{MyRed2}\textbf{.387}} & {\color{MyRed2}\textbf{.406}} & .423 & .418 & {\color{MyRed2}\textbf{.387}} & {\color{MyRed2}\textbf{.410}} & .416 & .433 & {\color{MyRed2}\textbf{.403}} & {\color{MyRed2}\textbf{.424}} & .420 & .425 & {\color{MyRed2}\textbf{.411}} & {\color{MyRed2}\textbf{.430}} & .428 & .431 \\
 \rowcolor{tabhighlight}\cellcolor{white}& AVG & {\color{MyRed2}\textbf{.361}} & {\color{MyRed2}\textbf{.397}} & .405 & .415 & {\color{MyRed2}\textbf{.356}} & {\color{MyRed2}\textbf{.391}} & .366 & .404  & {\color{MyRed2}\textbf{.349}} & {\color{MyRed2}\textbf{.370}} & .367 & .377  & {\color{MyRed2}\textbf{.346}} & {\color{MyRed2}\textbf{.372}} & .362  & .382 & {\color{MyRed2}\textbf{.355}} & {\color{MyRed2}\textbf{.377}} & .366 & .381  & {\color{MyRed2}\textbf{.358}} & {\color{MyRed2}\textbf{.382}} & .370 & .385 \\
\midrule
\multirow{5}{*}{\rotatebox{90}{ETTm1}} & 96 & .327 & {\color{MyRed2}\textbf{.343}} & {\color{MyRed2}\textbf{.309}} & .357 & {\color{MyRed2}\textbf{.334}} & {\color{MyRed2}\textbf{.350}} & .338 & .368 & {\color{MyRed2}\textbf{.355}} & {\color{MyRed2}\textbf{.355}} & .380 & .361 & {\color{MyRed2}\textbf{.348}} & {\color{MyRed2}\textbf{.354}} & .363 & .356 & {\color{MyRed2}\textbf{.310}} & {\color{MyRed2}\textbf{.327}} & .331 & .333 & {\color{MyRed2}\textbf{.314}} & {\color{MyRed2}\textbf{.331}} & .328 & .332 \\
 & 192 & .368 & {\color{MyRed2}\textbf{.378}} & {\color{MyRed2}\textbf{.346}} & .381 & .388 & {\color{MyRed2}\textbf{.386}} & {\color{MyRed2}\textbf{.353}} & .388 & {\color{MyRed2}\textbf{.388}} & {\color{MyRed2}\textbf{.380}} & .412 & .383 & {\color{MyRed2}\textbf{.385}} & .378 & .388 & {\color{MyRed2}\textbf{.375}} & {\color{MyRed2}\textbf{.363}} & {\color{MyRed2}\textbf{.360}} & .386 & .365 & {\color{MyRed2}\textbf{.364}} & {\color{MyRed2}\textbf{.365}} & .365 & .384 \\
 & 336 & {\color{MyRed2}\textbf{.373}} & {\color{MyRed2}\textbf{.396}} & {\color{MyRed2}\textbf{.373}} & .408 & .400 & {\color{MyRed2}\textbf{.412}} & {\color{MyRed2}\textbf{.381}} & .413 & {\color{MyRed2}\textbf{.399}} & {\color{MyRed2}\textbf{.387}} & .436 & .400 & {\color{MyRed2}\textbf{.410}} & .394 & .416 & {\color{MyRed2}\textbf{.392}} & .410 & .387 & {\color{MyRed2}\textbf{.408}} & {\color{MyRed2}\textbf{.382}} & {\color{MyRed2}\textbf{.391}} & {\color{MyRed2}\textbf{.417}} & {\color{MyRed2}\textbf{.391}} & .425 \\
 & 720 & {\color{MyRed2}\textbf{.445}} & {\color{MyRed2}\textbf{.438}} & .475 & .477 & {\color{MyRed2}\textbf{.457}} & {\color{MyRed2}\textbf{.451}} & .504 & .493 & {\color{MyRed2}\textbf{.429}} & {\color{MyRed2}\textbf{.413}} & .462 & .420 & {\color{MyRed2}\textbf{.448}} & {\color{MyRed2}\textbf{.416}} & .460 & .418 & {\color{MyRed2}\textbf{.477}} & {\color{MyRed2}\textbf{.427}} & .503 & .430 & .452 & {\color{MyRed2}\textbf{.521}} & {\color{MyRed2}\textbf{.445}} & .525 \\
 \rowcolor{tabhighlight}\cellcolor{white}& AVG & .378 & {\color{MyRed2}\textbf{.388}} & {\color{MyRed2}\textbf{.375}} & .405 & {\color{MyRed2}\textbf{.394}} & {\color{MyRed2}\textbf{.399}} & {\color{MyRed2}\textbf{.394}} & .415 & {\color{MyRed2}\textbf{.392}} & {\color{MyRed2}\textbf{.383}} & .422 & .391 & {\color{MyRed2}\textbf{.397}} & {\color{MyRed2}\textbf{.385}} & .406  & {\color{MyRed2}\textbf{.385}}  & {\color{MyRed2}\textbf{.390}} & {\color{MyRed2}\textbf{.375}} & .407 & .377 & {\color{MyRed2}\textbf{.380}} & {\color{MyRed2}\textbf{.408}} & .382 & .416 \\
\midrule
\multirow{5}{*}{\rotatebox{90}{ETTm2}} & 96 & {\color{MyRed2}\textbf{.180}} & {\color{MyRed2}\textbf{.259}} & .197 & .286 & {\color{MyRed2}\textbf{.181}} & {\color{MyRed2}\textbf{.260}} & .201 & .291 & {\color{MyRed2}\textbf{.192}} & {\color{MyRed2}\textbf{.259}} & .211 & .274 & {\color{MyRed2}\textbf{.194}} & {\color{MyRed2}\textbf{.265}} & .205 & .273 & {\color{MyRed2}\textbf{.175}} & .249 & .177 & {\color{MyRed2}\textbf{.244}} & {\color{MyRed2}\textbf{.180}} & {\color{MyRed2}\textbf{.248}} & {\color{MyRed2}\textbf{.180}} & .251 \\
 & 192 & {\color{MyRed2}\textbf{.245}} & {\color{MyRed2}\textbf{.305}} & .250 & .322 & {\color{MyRed2}\textbf{.247}} & {\color{MyRed2}\textbf{.307}} & .258 & .334 & {\color{MyRed2}\textbf{.256}} & {\color{MyRed2}\textbf{.302}} & .281 & .318 & {\color{MyRed2}\textbf{.257}} & {\color{MyRed2}\textbf{.304}} & .275 & .316 & {\color{MyRed2}\textbf{.242}} & {\color{MyRed2}\textbf{.290}} & .251 & .293 & {\color{MyRed2}\textbf{.243}} & {\color{MyRed2}\textbf{.292}} & .251 & .298 \\
 & 336 & {\color{MyRed2}\textbf{.283}} & {\color{MyRed2}\textbf{.338}} & .337 & .375 & {\color{MyRed2}\textbf{.293}} & {\color{MyRed2}\textbf{.344}} & .324 & .373 & {\color{MyRed2}\textbf{.289}} & {\color{MyRed2}\textbf{.329}} & .341 & .355 & {\color{MyRed2}\textbf{.301}} & {\color{MyRed2}\textbf{.342}} & .329 & .350 & {\color{MyRed2}\textbf{.299}} & {\color{MyRed2}\textbf{.326}} & .305 & .327 & {\color{MyRed2}\textbf{.302}} & {\color{MyRed2}\textbf{.331}} & .315 & .338 \\
 & 720 & {\color{MyRed2}\textbf{.364}} & {\color{MyRed2}\textbf{.392}} & .480 & .461 & {\color{MyRed2}\textbf{.376}} & {\color{MyRed2}\textbf{.396}} & .488 & .464 & {\color{MyRed2}\textbf{.372}} & {\color{MyRed2}\textbf{.384}} & .428 & .428 & {\color{MyRed2}\textbf{.387}} & {\color{MyRed2}\textbf{.396}} & .437 & .411 & {\color{MyRed2}\textbf{.394}} & {\color{MyRed2}\textbf{.387}} & .419 & .394 & {\color{MyRed2}\textbf{.406}} & {\color{MyRed2}\textbf{.396}} & .421 & .403 \\
 \rowcolor{tabhighlight}\cellcolor{white}& AVG & {\color{MyRed2}\textbf{.268}} & {\color{MyRed2}\textbf{.323}} & .316 & .361 & {\color{MyRed2}\textbf{.274}} & {\color{MyRed2}\textbf{.326}} & .317 & .365 & {\color{MyRed2}\textbf{.277}} & {\color{MyRed2}\textbf{.318}} & .315 & .343  & {\color{MyRed2}\textbf{.284}} & {\color{MyRed2}\textbf{.326}} & .311 & .337 & {\color{MyRed2}\textbf{.277}} & {\color{MyRed2}\textbf{.313}} & .288 & .314 & {\color{MyRed2}\textbf{.282}} & {\color{MyRed2}\textbf{.316}} & .291 & .330 \\
\midrule
\multirow{5}{*}{\rotatebox{90}{Weather}} & 96 & .161 & {\color{MyRed2}\textbf{.209}} & {\color{MyRed2}\textbf{.159}} & .213 & .163 & {\color{MyRed2}\textbf{.213}} & {\color{MyRed2}\textbf{.160}} & .214 & {\color{MyRed2}\textbf{.168}} & {\color{MyRed2}\textbf{.200}} & .278 & .376 & {\color{MyRed2}\textbf{.171}} & {\color{MyRed2}\textbf{.202}} & .220 & {\color{MyRed2}\textbf{.217}} & {\color{MyRed2}\textbf{.163}} & {\color{MyRed2}\textbf{.197}} & .177 & .210 & {\color{MyRed2}\textbf{.164}} & {\color{MyRed2}\textbf{.198}} & .172 & .206 \\
 & 192 & .217 & {\color{MyRed2}\textbf{.261}} & {\color{MyRed2}\textbf{.215}} & .266 & .215 & .263 & {\color{MyRed2}\textbf{.210}} & {\color{MyRed2}\textbf{.260}} & {\color{MyRed2}\textbf{.246}} & {\color{MyRed2}\textbf{.217}} & .301 & .409 & {\color{MyRed2}\textbf{.218}} & {\color{MyRed2}\textbf{.247}} & .271 & .259 & {\color{MyRed2}\textbf{.210}} & {\color{MyRed2}\textbf{.241}} & .224 & .253 & {\color{MyRed2}\textbf{.213}} & {\color{MyRed2}\textbf{.244}} & .218 & .248 \\
 & 336 & {\color{MyRed2}\textbf{.276}} & {\color{MyRed2}\textbf{.304}} & .291 & .322 & {\color{MyRed2}\textbf{.273}} & {\color{MyRed2}\textbf{.297}} & .309 & .309 & {\color{MyRed2}\textbf{.288}} & {\color{MyRed2}\textbf{.275}} & .329 & .420 & {\color{MyRed2}\textbf{.278}} & {\color{MyRed2}\textbf{.291}} & .286 & .297 & .264 & .282 & {\color{MyRed2}\textbf{.260}} & {\color{MyRed2}\textbf{.276}} & .273 & .288 & {\color{MyRed2}\textbf{.266}} & {\color{MyRed2}\textbf{.282}} \\
 & 720 & {\color{MyRed2}\textbf{.342}} & {\color{MyRed2}\textbf{.353}} & .415 & .400 & {\color{MyRed2}\textbf{.328}} & {\color{MyRed2}\textbf{.339}} & .418 & .405 & {\color{MyRed2}\textbf{.351}} & {\color{MyRed2}\textbf{.375}} & .370 & .463 & {\color{MyRed2}\textbf{.370}} & {\color{MyRed2}\textbf{.350}} & .373 & .354 & {\color{MyRed2}\textbf{.339}} & .334 & .345 & {\color{MyRed2}\textbf{.331}} & {\color{MyRed2}\textbf{.349}} & .342 & .358 & {\color{MyRed2}\textbf{.339}} \\
 \rowcolor{tabhighlight}\cellcolor{white}& AVG & {\color{MyRed2}\textbf{.249}} & {\color{MyRed2}\textbf{.281}} & .270 & .300 & {\color{MyRed2}\textbf{.244}} & {\color{MyRed2}\textbf{.278}} & .274 & .297 & {\color{MyRed2}\textbf{.263}} & {\color{MyRed2}\textbf{.266}} & .319 & .417 & {\color{MyRed2}\textbf{.259}} & {\color{MyRed2}\textbf{.272}} & .287 & .281  & {\color{MyRed2}\textbf{.244}} & {\color{MyRed2}\textbf{.263}} & .251 & .267 & {\color{MyRed2}\textbf{.249}} & {\color{MyRed2}\textbf{.268}} & .253 & {\color{MyRed2}\textbf{.268}} \\
\midrule
\multirow{5}{*}{\rotatebox{90}{GlobalTemp}} & 96 & .226 & .346 & {\color{MyRed2}\textbf{.219}} & {\color{MyRed2}\textbf{.341}} & {\color{MyRed2}\textbf{.229}} & {\color{MyRed2}\textbf{.349}} & .230 & .350 & {\color{MyRed2}\textbf{.234}} & {\color{MyRed2}\textbf{.351}} & .278 & .376 & {\color{MyRed2}\textbf{.236}} & {\color{MyRed2}\textbf{.352}} & .273 & .377 & {\color{MyRed2}\textbf{.226}} & {\color{MyRed2}\textbf{.343}} & .236 & .352 & {\color{MyRed2}\textbf{.230}} & {\color{MyRed2}\textbf{.345}} & .233 & .348 \\
 & 192 & {\color{MyRed2}\textbf{.263}} & .386 & .265 & {\color{MyRed2}\textbf{.381}} & .272 & .390 & {\color{MyRed2}\textbf{.268}} & {\color{MyRed2}\textbf{.385}} & {\color{MyRed2}\textbf{.266}} & {\color{MyRed2}\textbf{.382}} & .301 & .409 & {\color{MyRed2}\textbf{.268}} & {\color{MyRed2}\textbf{.384}} & .304 & .409 & {\color{MyRed2}\textbf{.271}} & {\color{MyRed2}\textbf{.384}} & .287 & .398 & {\color{MyRed2}\textbf{.280}} & {\color{MyRed2}\textbf{.389}} & .287 & .397 \\
 & 336 & {\color{MyRed2}\textbf{.309}} & {\color{MyRed2}\textbf{.420}} & .326 & .426 & {\color{MyRed2}\textbf{.311}} & {\color{MyRed2}\textbf{.423}} & .326 & .427 & {\color{MyRed2}\textbf{.309}} & {\color{MyRed2}\textbf{.420}} & .329 & {\color{MyRed2}\textbf{.420}} & {\color{MyRed2}\textbf{.309}} & {\color{MyRed2}\textbf{.420}} & .332 & .437 & {\color{MyRed2}\textbf{.314}} & {\color{MyRed2}\textbf{.419}} & .332 & .433 & {\color{MyRed2}\textbf{.318}} & {\color{MyRed2}\textbf{.420}} & .320 & .430 \\
 & 720 & {\color{MyRed2}\textbf{.340}} & {\color{MyRed2}\textbf{.447}} & .344 & .453 & {\color{MyRed2}\textbf{.343}} & {\color{MyRed2}\textbf{.449}} & .377 & .467 & {\color{MyRed2}\textbf{.361}} & {\color{MyRed2}\textbf{.459}} & .379 & .467 & {\color{MyRed2}\textbf{.347}} & {\color{MyRed2}\textbf{.449}} & .379 & .469 & {\color{MyRed2}\textbf{.427}} & {\color{MyRed2}\textbf{.502}} & .463 & .524 & {\color{MyRed2}\textbf{.438}} & {\color{MyRed2}\textbf{.504}} & .452 & .521 \\
 &\cellcolor{tabhighlight} AVG &\cellcolor{tabhighlight} {\color{MyRed2}\textbf{.284}} &\cellcolor{tabhighlight} {\color{MyRed2}\textbf{.399}} &\cellcolor{tabhighlight} .288 &\cellcolor{tabhighlight} .400 &\cellcolor{tabhighlight} {\color{MyRed2}\textbf{.288}} &\cellcolor{tabhighlight} {\color{MyRed2}\textbf{.402}} &\cellcolor{tabhighlight} .300 &\cellcolor{tabhighlight} .407  &\cellcolor{tabhighlight} {\color{MyRed2}\textbf{.292}} &\cellcolor{tabhighlight} {\color{MyRed2}\textbf{.403}} &\cellcolor{tabhighlight} .321 &\cellcolor{tabhighlight} .418 &\cellcolor{tabhighlight} {\color{MyRed2}\textbf{.290}} &\cellcolor{tabhighlight} {\color{MyRed2}\textbf{.401}} &\cellcolor{tabhighlight} .322 &\cellcolor{tabhighlight} .423 &\cellcolor{tabhighlight} {\color{MyRed2}\textbf{.322}} &\cellcolor{tabhighlight} {\color{MyRed2}\textbf{.418}} &\cellcolor{tabhighlight} .329 &\cellcolor{tabhighlight} .426  &\cellcolor{tabhighlight} {\color{MyRed2}\textbf{.316}} &\cellcolor{tabhighlight} {\color{MyRed2}\textbf{.414}} &\cellcolor{tabhighlight} .323 &\cellcolor{tabhighlight} .424 \\
\midrule
\rowc\rowcolor{blue!15}
\multicolumn{2}{c|}{\scalebox{1.1}{\textbf{\# Wins}}} & \scalebox{1.1}{\color{MyRed2}\textbf{22}} & \scalebox{1.1}{\color{MyRed2}\textbf{28}} & 9 & 2 & \scalebox{1.1}{\color{MyRed2}\textbf{23}} & \scalebox{1.1}{\color{MyRed2}\textbf{28}} & 9 & 2 & \scalebox{1.1}{\color{MyRed2}\textbf{30}} & \scalebox{1.1}{\color{MyRed2}\textbf{30}} & 0 & 0 & \scalebox{1.1}{\color{MyRed2}\textbf{30}} & \scalebox{1.1}{\color{MyRed2}\textbf{28}} &0 & 3 & \scalebox{1.1}{\color{MyRed2}\textbf{28}} & \scalebox{1.1}{\color{MyRed2}\textbf{26}} & 2 & 5 & \scalebox{1.1}{\color{MyRed2}\textbf{28}} & \scalebox{1.1}{\color{MyRed2}\textbf{28}} & 4 & 3 \\
\bottomrule[1.5pt]
\end{tabular}}
\label{tab:main_result}
\end{table*}

%% file: tables/gift_eval.tex
\begin{table*}[]
\setlength{\abovecaptionskip}{.2 cm}
\caption{
Performance comparison on the GIFT-Eval benchmark.
Data previously included in Time-300B, LOTSA, and BLAST have been excluded.
Lower MASE values indicate better performance.
Models with superior performance are highlighted in {\color{MyRed2}\textbf{red}}.
}
\setlength\tabcolsep{1.2pt}
\resizebox{1\linewidth}{!}{
\begin{tabular}{rr|cc|cc|cc|cc|cc|cc}
\toprule[1.5pt]
\multicolumn{2}{l}{\textbf{Models}} & \multicolumn{1}{|l}{\textbf{TimeMoE}$_{l}^\dagger$} & \multicolumn{1}{l|}{\textbf{TimeMoE}$_{l}$} & \multicolumn{1}{l}{\textbf{TimeMoE}$_{b}^\dagger$} & \multicolumn{1}{l|}{\textbf{TimeMoE}$_{b}$} & \multicolumn{1}{l}{\textbf{MOIRAI}$_{l}^\dagger$} & \multicolumn{1}{l|}{\textbf{MOIRAI}$_{l}$} & \multicolumn{1}{l}{\textbf{MOIRAI}$_{b}^\dagger$} & \multicolumn{1}{l|}{\textbf{MOIRAI}$_{b}$} & \multicolumn{1}{l}{\textbf{Chronos}$_{b}^\dagger$} & \multicolumn{1}{l|}{\textbf{Chronos}$_{b}$} & \multicolumn{1}{l}{\textbf{Chronos}$_{s}^{\dagger}$} & \multicolumn{1}{l}{\textbf{Chronos}$_{s}$} \\
\midrule
\multicolumn{2}{c|}{\textbf{MASE}} &  \scalebox{1.1}{\color{MyRed2}\textbf{0.777}} & 0.872 & \scalebox{1.1}{\color{MyRed2}\textbf{0.760}} & 0.888 & \scalebox{1.1}{\color{MyRed2}\textbf{0.740}} & 0.816 & \scalebox{1.1}{\color{MyRed2}\textbf{0.759}} & 0.812 & \scalebox{1.1}{\color{MyRed2}\textbf{0.711}} & 0.740 & \scalebox{1.1}{\color{MyRed2}\textbf{0.738}} & 0.742 \\
\bottomrule[1.5pt]
\end{tabular}}
\label{tab:gift_eval}
\end{table*}

%% file: tables/ablation.tex
\begin{table*}[t]
\setlength{\abovecaptionskip}{0.2 cm}
\caption{Performance of different sampling strategies. 
Models with superior performance are highlighted in {\color{MyRed2}\textbf{red}}.
}
\setlength\tabcolsep{2.5pt}
\resizebox{1\linewidth}{!}{
\begin{tabular}{c|cccc|cccc|cccc|cccc|cccc}
\toprule[1.5pt]
\multicolumn{1}{c}{} & \multicolumn{4}{c}{\textbf{Naive Sampling}} & \multicolumn{4}{c}{\textbf{Stratified Sampling}}  & \multicolumn{4}{c}{\textbf{Balanced Sampling}} & \multicolumn{4}{c}{\textbf{\textit{w/o GS}}} & \multicolumn{4}{c}{\textbf{\textit{w/o GM}}}  \\
\cmidrule(r){2-5} \cmidrule(r){6-9} \cmidrule(r){10-21}
\textbf{Horizons} & \textbf{96} & \textbf{192} & \textbf{336} & \textbf{720} & \textbf{96} & \textbf{192} & \textbf{336} & \textbf{720}& \textbf{96} & \textbf{192} & \textbf{336} & \textbf{720}& \textbf{96} & \textbf{192} & \textbf{336} & \textbf{720}& \textbf{96} & \textbf{192} & \textbf{336} & \textbf{720}    \\
\midrule
ETTh1  &0.393&0.421&0.468&0.507&0.388&0.412&0.458&0.503&{\color{MyRed2}\textbf{0.376}}&{\color{MyRed2}\textbf{0.401}}&{\color{MyRed2}\textbf{0.419}}&{\color{MyRed2}\textbf{0.455}}&0.390&0.418&0.448&0.489&0.386&0.414&0.444&0.483\\
\midrule
ETTh2  &0.364&0.401&0.460&0.513&0.362&0.399&0.458&0.495&{\color{MyRed2}\textbf{0.332}}&{\color{MyRed2}\textbf{0.378}}&{\color{MyRed2}\textbf{0.405}}&{\color{MyRed2}\textbf{0.452}}&0.366&0.401&0.455&0.500&0.338&0.388&0.422&0.465\\
\midrule
ETTm1  &0.379&0.422&0.454&0.501&0.372&0.413&0.450&0.494&{\color{MyRed2}\textbf{0.350}}&{\color{MyRed2}\textbf{0.386}}&{\color{MyRed2}\textbf{0.412}}&{\color{MyRed2}\textbf{0.451}}&0.366&0.412&0.450&0.485&0.355&0.389&0.435&0.474\\
\midrule
ETTm2  &0.303&0.344&0.381&0.419&0.299&0.330&0.376&0.409&{\color{MyRed2}\textbf{0.260}}&{\color{MyRed2}\textbf{0.307}}&{\color{MyRed2}\textbf{0.344}}&{\color{MyRed2}\textbf{0.396}}&0.305&0.338&0.370&0.403&0.275&0.318&0.353&0.400    \\
\bottomrule[1.5pt]
\end{tabular}
}
\label{ablation:tab}
\end{table*}

%% file: sections/conclusion.tex
\section{Conclusion}
In this work, we present BLAST, a balanced sampling time series corpus designed to address the critical yet understudied challenge of data diversity in training universal forecasting models.
By integrating 321 billion observations from diverse public datasets and introducing a novel balanced sampling strategy, BLAST systematically mitigates inherent biases in large-scale time series distributions.
The proposed balanced sampling techniques ensure representative pattern coverage, thereby enhancing both the training efficiency and generalization capability of the model.
Extensive experiments demonstrate that models pre-trained on BLAST achieve superior zero-shot forecasting accuracy, outperforming models trained on naively or stratified-sampled corpora.

\begin{acks}
This study is partially supported by NSFC No. 62372430, the Youth Innovation Promotion Association CAS No.2023112, and HUA-Innovation fundings.
We also gratefully acknowledge the reviewers for their constructive comments and thorough discussions, which were instrumental in improving the quality of this manuscript.
\end{acks}

%% file: sections/appendix.tex
\begin{table*}[t]
  \centering
  \setlength{\abovecaptionskip}{0.cm}
  \setlength{\belowcaptionskip}{0cm}
  \caption{Hyperparameter study for \texttt{n\_neighbor}.}
  \includegraphics[width=0.92\linewidth]{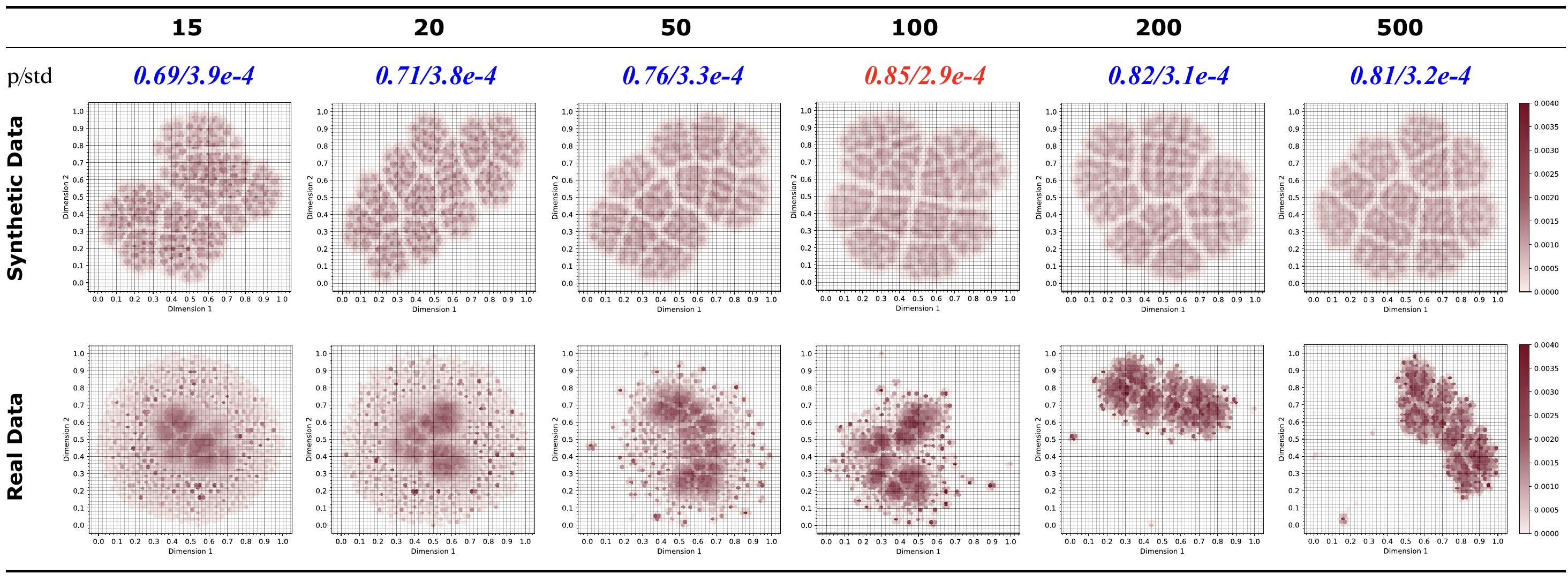}
  \label{umap_n_neighbor}
\end{table*}

\begin{table*}[t]
  \centering
  \setlength{\abovecaptionskip}{0.cm}
  \setlength{\belowcaptionskip}{-0.4cm}
  \caption{Hyperparameter study for \texttt{min\_dist}.}
  \includegraphics[width=0.92\linewidth]{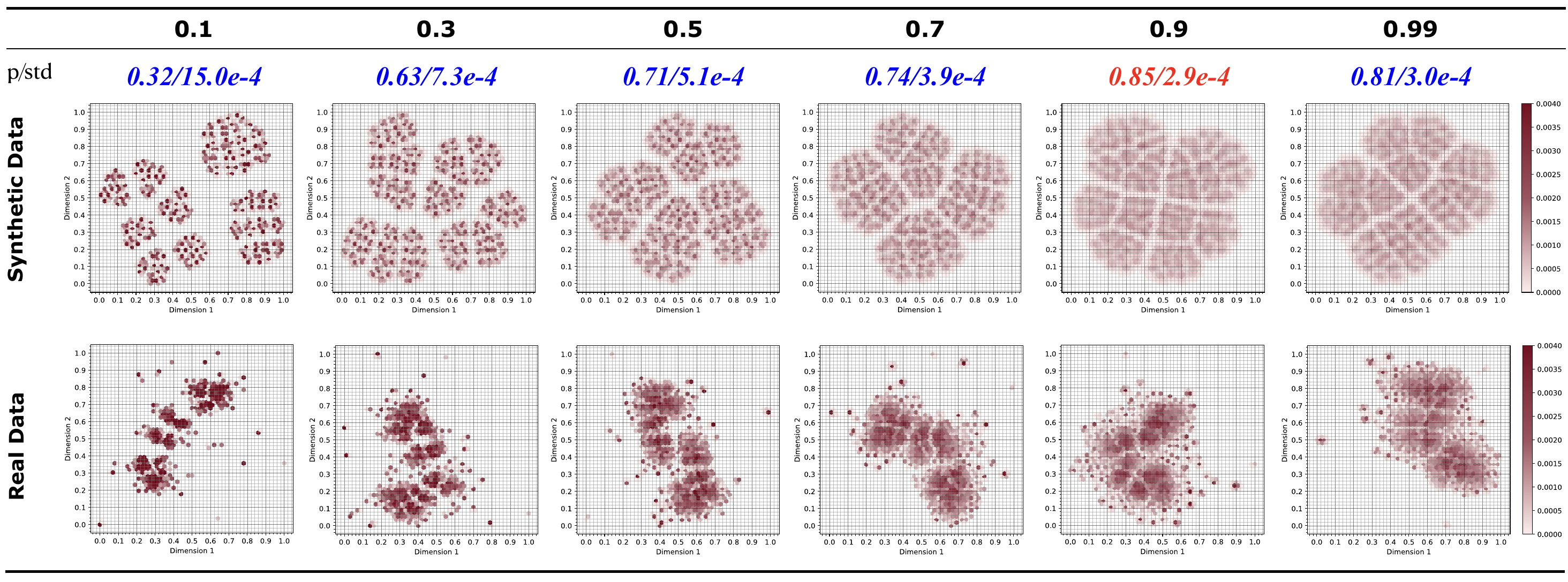}
  \label{umap_min_dist}
\end{table*}

\appendix

\section{Details of BLAST}

\label{appendix:A}
\subsection{Raw Data Construction}
\label{appendix:A.1}
Building upon previous work~\cite{MOMENT, Chronos, MOIRAI, UAD, UCR}, we have collected a large-scale time series dataset reaching 321 billion data points.
It is important to note that not all data were used for training.
Common benchmark datasets, such as ETT, Weather, and Traffic, were excluded to ensure the integrity of our experimental settings. Additionally, we filtered out time series with more than 5\% missing values (NaN).
Moreover, we retained the remaining NaN values within the filtered time series.
These missing values are handled dynamically during the training phase, according to the specific requirements of the model.

\subsection{Metrics Calculation}
\label{appendix:A.2}
\subsubsection{\textbf{Selection Principles for Metrics}}
Metrics selection is essential for effectively capturing the underlying patterns of a time series.
The seven metrics selected in this study are widely used in statistical time series analysis, and each highlight different dynamic aspects, providing a comprehensive and complementing representation of the series' pattern.
For example, trends and seasonality capture distinct components: trends represent low-frequency, long-term variations, while seasonality reflects high-frequency, periodic fluctuations.
Stability, volatility, hetero/homo-scedasticity, and anomalies present distributional characteristics and variability from different angles.
Furthermore, the combination of memory and seasonality could reveal the long-term dependency structure within the data.
Additionally, it is crucial that \textit{these metrics should not be directly tied to predictability}; otherwise, harmful samples may be introduced during the grid sampling process.

\input{tables/full_shot}

\subsubsection{\textbf{Handling Variable-Length Series}}
Although these metrics do not have stringent requirements on time series length, excessively long samples may result in less robust representations.
Therefore, we standardize time series to a maximum context length of 4096.
Specifically, for time series longer than 4096, we randomly sample three segments and compute the metrics for each.
For continuous metrics, we take the average, while for discrete metrics, we use a voting strategy to select the most frequent value.

\subsubsection{\textbf{Alternative Methods Considered}}
The core objective of metrics calculation is to comprehensively capture the patterns of a time series.
Any method capable of achieving this goal can be applied at this stage.
One potential alternative is using deep learning models to generate time series representations.
However, the raw BLAST dataset is vast, containing 40 million time series, and there is currently no widely recognized and robust model for time series representation that can process such large-scale data efficiently.
Additionally, while using statistical metrics to characterize a time series is significantly faster than deep learning models, it still requires considerable time. In our experiments, this process took 8 days using 128 CPU threads (Intel Xeon 6338 2.0GHz). Therefore, improving the efficiency of time series representation in the balanced sampling process remains a critical topic for future research.

\subsection{UMAP Hyperparameter Study}
\label{appendix:A.3}

\subsubsection{\textbf{UMAP Hyperparameter Description}}
The choice of UMAP parameters significantly impacts dimension reduction. In this study, the primary goal is to preserve the global structure of the large-scale dataset, particularly its overall distribution. Key UMAP parameters include \texttt{n\_neighbors}, \texttt{min\_dist}, and \texttt{metric}, which influence different aspects of the embedding process.

\begin{itemize}
    \item \textbf{\texttt{n\_neighbors}}: This parameter controls the balance between local and global structures. Larger values better capture the global distribution by considering more neighbors.
    \item \textbf{\texttt{min\_dist}}: This determines the compactness of points in the reduced space. A higher value prevents excessive clustering of points, prioritizing the preservation of global topology.
    \item \textbf{\texttt{metric}}: This defines the distance function for measuring point similarity. Given the discretization process in BLAST, we use the Hamming distance, calculated as $d_H(x, y) = \sum_{i} \mathbb{I}(x_i \neq y_i)$, where \( x \) and \( y \) are two feature vectors, and \( \mathbb{I}(\cdot) \) is an indicator function that returns 1 if the condition is true and 0 otherwise.
\end{itemize}

\subsubsection{\textbf{Hyperparameter Optimization}}
Similar to $\S$ \ref{sec:6_5}, to optimize UMAP parameters, we generated synthetic data with uniformly distributed feature vectors, following the feature construction process in BLAST. Specifically, each one-hot feature was uniformly assigned across categories. 

We assessed parameter effectiveness using two metrics: the \textbf{proportion of non-empty grids} (\( \text{p} \)) and the \textbf{standard deviation of grid density} (\( \text{std} \)) after dimension reduction.
Larger $\text{p}$ and smaller $\text{std}$ indicate better results.
Using \texttt{n\_neighbors} = 100 and \texttt{min\_dist} = 0.9 as the baseline, we tested values for \texttt{n\_neighbors} in the range \([15, 20, 50, 100, 200, 500]\) and for \texttt{min\_dist} in \([0.1, 0.3, 0.5, 0.7, 0.9,\\ 0.99]\).
The results, shown in Table~\ref{umap_n_neighbor} and Table~\ref{umap_min_dist}, show that the dimension reduction is optimal when \texttt{n\_neighbors} = 100 and \texttt{min\_dist} = 0.9, with consistent trends observed for both real and synthetic datasets.

\subsection{Using the BLAST Corpus}
To facilitate user access, we directly provide the processed data.
These datasets are represented as \( N \times L \) matrices, where \( N \) denotes the number of samples, and \( L \) is the length of each sample.
The length \( L \) is set to 4096, and samples shorter than 4096 are right-padded with NaN values to ensure uniform length. This approach allows users to easily read and utilize the samples.






\section{Details of Experiments}
\label{appendix:B}

\subsection{Evaluation Metrics}
In this study, we use the Mean Absolute Error (MAE) and Mean Squared Error (MSE) as evaluation metrics. These metrics are commonly used to assess the performance of predictive models and can be formulated as follows:
\begin{equation}
    \text{MAE} = \frac{1}{N} \sum_{i=1}^{N} \left| y_i - \hat{y}_i \right|,\qquad \text{MSE} = \frac{1}{N} \sum_{i=1}^{N} \left( y_i - \hat{y}_i \right)^2
\end{equation}
where \( y_i \) is the true value, \( \hat{y}_i \) is the predicted value, and \( N \) is the total number of samples.

\subsection{Details for Benchmark Datasets.}
The ETTh1, ETTh2, ETTm1, ETTm2, and Weather datasets adhere to the standard settings established in previous studies. For evaluation, we utilize the test set for zero-shot prediction. The results we obtained are consistent with those reported in TimeMoE~\cite{TimeMoE}.  
For the GlobalWeather dataset, since TimeMoE~\cite{TimeMoE} does not follow conventional settings and lacks detailed descriptions, we perform the evaluation using the test set~\cite{Corrformer}, applying z-score normalization and setting the stride \( S \) equal to the prediction length.
For the GIFT-Eval benchmark, we follow its original setting.

\subsection{Additional Results}

We compare TimeMoE pretrained on BLAST with full-shot models~\cite{iTransformer, TimesNet, PatchTST, DLinear, deng2024disentangling, deng2024parsimony} on the ETTh1, ETTh2, ETTm1, ETTm2, and Weather datasets. Following the experimental settings in TimeMoE~\cite{TimeMoE}, we report the average error in Table~\ref{full_shot}. It can be observed that BLAST-pretrained TimeMoE outperforms these full-shot models.


%% file: tables/full_shot.tex
\begin{table*}[t]
\setlength{\abovecaptionskip}{0.cm}
\caption{
Performance comparison of TimeMoE pre-trained on BLAST against full-shot models. {\color{red}\textbf{Red}}: the best, {\color{blue}Blue}: 2nd best.}
\resizebox{0.93\linewidth}{!}{
\begin{tabular}{rr|cc|cc|cc|cc|cc|cc|cc|cc}
\toprule[1.5pt]
\multicolumn{2}{r}{\multirow{2}{*}{\textbf{Models}}} & \multicolumn{4}{c}{\textbf{Pre-training on BLAST}} & \multicolumn{12}{c}{\textbf{Full-shot Models}}\\
\cmidrule(r){3-6} \cmidrule(r){7-18}
\multicolumn{2}{r}{}            & \multicolumn{2}{|c}{\textbf{TimeMoE}$_{base}$} & \multicolumn{2}{c|}{\textbf{TimeMoE}$_{large}$} & \multicolumn{2}{c|}{\textbf{iTransformer}} & \multicolumn{2}{c|}{\textbf{TimeMixer}} & \multicolumn{2}{c|}{\textbf{TimesNet}} & \multicolumn{2}{c|}{\textbf{PatchTST}} & \multicolumn{2}{c|}{\textbf{TiDE}}& \multicolumn{2}{c}{\textbf{DLinear}} \\
\cmidrule(r){3-4} \cmidrule(r){5-6} \cmidrule(r){7-8} \cmidrule(r){9-10} \cmidrule(r){11-12} \cmidrule(r){13-14} \cmidrule(r){15-16} \cmidrule(r){17-18}
\multicolumn{2}{r|}{\textbf{Metrics}} & \textbf{MSE} & \textbf{MAE} & \textbf{MSE} & \textbf{MAE} & \textbf{MSE} & \textbf{MAE} & \textbf{MSE}  & \textbf{MAE} & \textbf{MSE} & \textbf{MAE} & \textbf{MSE}  & \textbf{MAE} & \textbf{MSE} & \textbf{MAE} & \textbf{MSE} & \textbf{MAE} \\
\midrule
\multicolumn{2}{r|}{\rotatebox{0}{ETTh1}} 
&\color{blue}\textbf{0.399}&\color{red}\textbf{0.412}&\color{red}\textbf{0.396}&\color{red}\textbf{0.412}&0.454&0.447&0.448&0.442&0.454&0.450&0.468&0.454&0.540&0.507&0.455&0.451\\
\midrule
\multicolumn{2}{r|}{\rotatebox{0}{ETTh2}} 
&\color{red}\textbf{0.356}&\color{red}\textbf{0.391}&\color{blue}\textbf{0.361}&0.397&0.383&0.406&0.364&\color{blue}\textbf{0.395}&0.414&0.496&0.386&0.406&0.611&0.549&0.558&0.515\\
\midrule
\multicolumn{2}{r|}{\rotatebox{0}{ETTm1}} 
&0.394&0.399&\color{red}\textbf{0.378}&\color{red}\textbf{0.388}&0.407&0.409&\color{blue}\textbf{0.381}&\color{blue}\textbf{0.395}&0.400&0.405&0.387&0.400&0.419&0.419&0.403&0.406\\
\midrule
\multicolumn{2}{r|}{\rotatebox{0}{ETTm2}} 
&\color{blue}\textbf{0.274}&\color{blue}\textbf{0.326}&\color{red}\textbf{0.268}&\color{red}\textbf{0.323}&0.288&0.332&0.275&\color{red}\textbf{0.323}&0.291&0.332&0.280&\color{blue}\textbf{0.326}&0.358&0.403&0.350&0.400\\
\midrule
\multicolumn{2}{r|}{\rotatebox{0}{Weather}} 
&\color{blue}\textbf{0.244}&\color{blue}\textbf{0.278}&0.249&0.281&0.257&\color{blue}\textbf{0.278}&\color{red}\textbf{0.240}&\color{red}\textbf{0.271}&0.258&0.286&0.258&0.280&0.270&0.320&0.265&0.316\\
\midrule
\rowc\rowcolor{blue!15}
\multicolumn{2}{c|}{\scalebox{1.1}{\textbf{Average}}} &{\color{blue}\textbf{0.333}}&{\color{blue}\textbf{0.361}}&{\color{red}\textbf{0.330}}&{\color{red}\textbf{0.36}}&0.357&0.374&0.341&0.365&0.363&0.393&0.355&0.373&0.439&0.439&0.406&0.417\\
\bottomrule[1.5pt]
\end{tabular}}
\label{full_shot}
\end{table*}